\documentclass{article}

\PassOptionsToPackage{numbers}{natbib}
\usepackage[main,final]{neurips_2026}
\usepackage[utf8]{inputenc}
\usepackage[T1]{fontenc}
\usepackage{hyperref}
\usepackage{url}
\usepackage{booktabs}
\usepackage{amsfonts}
\usepackage{amsmath,amssymb,amsthm,mathtools}
\usepackage{bm}
\usepackage{nicefrac}
\usepackage{microtype}
\usepackage{xcolor}
\usepackage{algorithm}
\usepackage{algorithmic}
\usepackage{enumitem}
\usepackage{booktabs, multirow}
\usepackage{subcaption}
\theoremstyle{plain}
\newtheorem{proposition}{Proposition}

\theoremstyle{remark}

\newcommand{\R}{\mathbb{R}}
\newcommand{\E}{\mathbb{E}}
\newcommand{\KL}{\mathrm{KL}}
\newcommand{\Gam}{\mathrm{Gamma}}
\newcommand{\N}{\mathcal{N}}
\newcommand{\St}{\mathcal{T}}
\newcommand{\bz}{\bm{z}}
\newcommand{\bx}{\bm{x}}
\newcommand{\bmu}{\bm{\mu}}
\newcommand{\dif}{\mathrm{d}}
\newcommand{\tr}{\mathrm{tr}}

\usepackage{xcolor} 
\newif\ifshowchanges
\showchangesfalse    
\newcommand{\rev}[1]{%
  \ifshowchanges{\color{blue}#1}\else#1\fi
}
\newcommand{\revr}[1]{%
  \ifshowchanges{\color{red}#1}\else#1\fi
}

\title{Pro-\texorpdfstring{$t$}{t}-VaDE: Robust Prototype Discovery via Closed-Form Student-\texorpdfstring{$t$}{t} Variational Deep Embedding}

\author{
  \textbf{Zhikang Xu}$^{1}$, \textbf{Jiarui Xing}$^{2}$, \textbf{Jian Wang}$^{3}$\thanks{\revr{The theoretical framework of this work was developed while the author was at Harvard Medical School; the author is currently affiliated with Eli Lilly and Company.}} \\
  $^1$Institute of Intelligent Information Processing, Shanxi University \\
  $^2$School of Medicine, Yale University \\
  $^3$Boston Children's Hospital, Harvard Medical School  \\
  \texttt{xuzhikangnba1@sxu.edu.cn}, \texttt{jiarui.xing@yale.edu}, \texttt{jianbljh@gmail.com} \\
}
\begin{document}
\maketitle


\begin{abstract}
We present a robust \revr{deep generative framework for unsupervised image prototype discovery, built on a heavy-tailed Student's $t$ mixture prior, that} bridges statistical robustness and mathematical tractability. Existing generative approaches typically rely on noise-prone Gaussian priors or computationally intensive diffusion models; neither simultaneously provides explicit latent structure nor exact analytic optimization. While diffusion models can yield strong priors, iterative sampling and intractable likelihoods limit scalability and interpretability for \revr{prototype discovery}. We address these issues by constructing a Student's $t$ latent space via a Bayesian scale-mixture formulation. This yields a fully closed-form augmented ELBO for \revr{unsupervised Student's $t$ mixture variational autoencoders, which, to our knowledge, is the first such objective applied to robust image prototype discovery}, avoiding biased variational approximations and high computational costs in prior heavy-tailed or diffusion-based frameworks. Treating precision as a latent variable automatically modulates each data point's influence, downweighting outliers to learn a data-dependent measure of trust. The result is improved robustness\revr{, accuracy, and efficiency, with well-separated latent representations} and more stable manifold learning than state-of-the-art methods. We validate our model on standard vision benchmarks and challenging neuroimaging tasks. In brain MRI-driven neurodegenerative disease analysis, our model recovers clean, anatomically coherent \revr{prototypes} where clinical noise often collapses rare subtypes. These results highlight the ability of our framework to \rev{preserve anatomy-associated patterns} for precise discovery in clinical settings. Our code is available at: \url{https://github.com/xuzhikangnba/torch-tvae}.
\end{abstract}

\section{Introduction}
\begin{figure}[!ht]
    \centering
    \includegraphics[width=\linewidth]{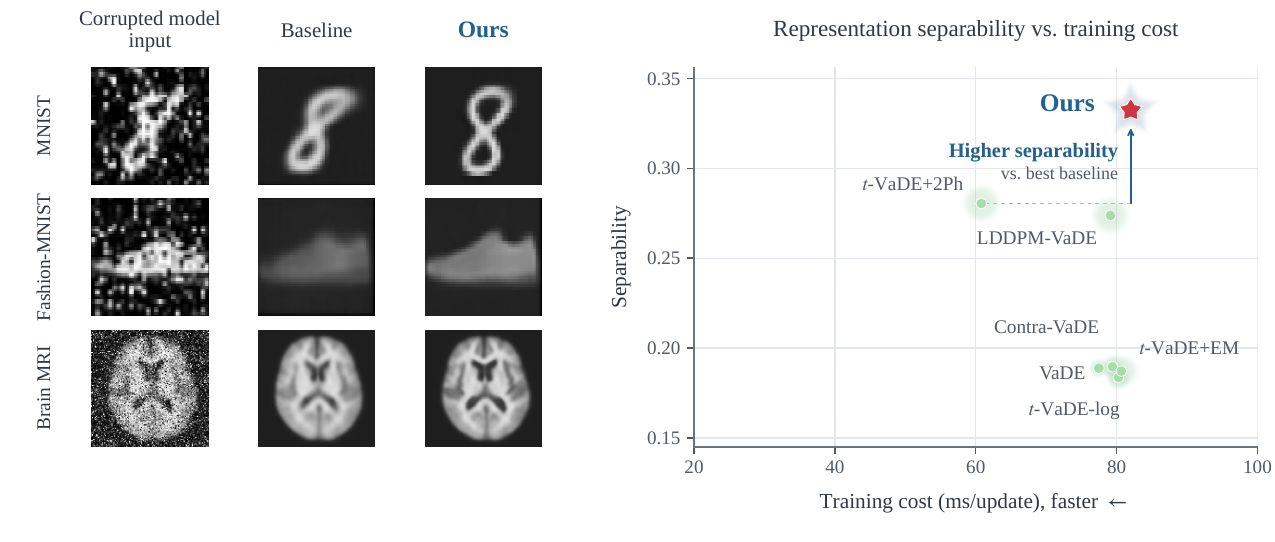}
    \caption{\textbf{Prototype generation and representation separability versus training cost.} \textit{Left:} Noise-corrupted inputs, diffusion-based prototype estimates, and our method on MNIST, Fashion-MNIST, and brain MRI. \textit{Right:} Mean class separability vs. per-update training time on brain scans; shading shows population variance.}
    \label{fig:teaser}
\end{figure}

A core goal in unsupervised visual representation learning is to compress large image collections into a small set of \textit{prototypes}: canonical exemplars that summarize data modes and support comparison, retrieval, and interpretation. Their utility hinges on structural integrity \cite{xie2016unsupervised, caron2018deep}: prototypes that capture coherent modes (not averaged blur) reveal latent structure that pooled descriptors can erase. This idea appears as cluster centroids in deep clustering \citep{xie2016unsupervised, caron2018deep,xing2026manifold}, object-centric representations in self-supervision \citep{caron2020unsupervised, caron2021emerging}, deformable atlases in medical imaging \citep{blaiotta2018generative, rakic2025atlasmorph,dalca2019learning}, and morphological exemplars in computational pathology \citep{song2024morphological}. Deformable atlas building methods \citep{rakic2025atlasmorph,wang2022geo} perform well when images share a consistent topology, but they frequently degrade under topological discrepancies (e.g., distinct object categories) or severe image artifacts, mainly due to their reliance on a globally defined template. In this work, we address this fundamental constraint by modeling data variation via flexible, probabilistic \revr{prototypes, defined as the component centers of a heavy-tailed generative mixture} in a learned latent space, providing intrinsic robustness to topological heterogeneity and distributional perturbations without requiring an explicit template.

Two paradigms dominate \revr{deep prototype learning, largely from deep clustering}. \textit{Discriminative/contrastive} methods jointly learn representations and cluster assignments, from early deep embedded clustering to modern self-supervised frameworks \cite{xie2016unsupervised,caron2021emerging,li2021contrastive,van2020scan,guo2017improved}. However, their prototypes are usually $k$-means centroids, lacking an explicit probabilistic treatment of uncertainty and noise. \textit{Generative \revr{mixture modeling}}, instead, embeds an explicit latent-space mixture in a deep generative model, where each prototype is the mean of an interpretable component density \cite{jiang2017variational, kim2024t3vae}. These methods mainly differ in objective tractability. Diffusion-based approaches either pair DDPMs with discrete clustering via EM-style updates \cite{ho2020denoising, yan2023clusterddpm, dasilva2024treediffusion,wang2023infodiffusion} or use diffusion as a generative regularizer \cite{hu2024gcc}; in both cases, per-sample losses are typically stochastic, and clustering is often applied post hoc rather than jointly optimized. VaDE \cite{jiang2017variational} is a key exception: its ELBO is fully closed form, enabling exact gradients and stable end-to-end training.

In VaDE-based frameworks, preserving analytical tractability while achieving statistical robustness is difficult. The standard Gaussian mixture prior penalizes distant latent codes quadratically, so even a few outliers can induce large gradients that pull \revr{prototypes} toward spurious modes and weaken cluster separation. Robust statistics therefore motivates heavy-tailed Student’s $t$ components to downweight extremes~\citep{lange1989robust,peel2000robust}. In deep generative models, however, a $t$ prior typically removes a closed-form ELBO. Prior $t$-based extensions~\citep{kobayashi2019vade,abiri2020variational} regain tractability via hierarchical variational bounds with hard-to-measure surrogate bias, while other heavy-tailed VAEs~\citep{kim2024t3vae} use high-variance Monte Carlo gradients. \revr{Although an exact closed-form $t$-mixture was derived for supervised authorship attribution by conditioning components on class labels~\citep{boenninghoff2020variational}, its extension to unsupervised image prototype discovery and robustness to heavy-tailed corruption are still untested.}

We close this gap by expressing the Student’s $t$ prior in its classical scale-mix form \cite{andrews1974scale}: a Student’s $t$ variable is a Gaussian whose precision is modulated by a Gamma latent variable\revr{, i.e., a Gaussian-Gamma hierarchy that we previously used for subject-specific regularization in Bayesian atlas building~\cite{wang2021bayesian}}. Treating precision as an explicit per-cluster latent variable and inferring it jointly with cluster assignments yields a strict generalization of VaDE. The marginal prior over cluster means remains Student's $t$ for heavy-tailed robustness, while conditionally it factorizes into Gaussian and Gamma terms that integrate out in closed form via conjugacy. This gives a theoretically principled\revr{, robust, and efficient generative framework for image prototype discovery and robust representation learning}. \rev{As shown in Figure~\ref{fig:teaser}, our method substantially improves representation separability with only a modest increase in per-update training time.} We summarize our core contributions as follows:

\begin{itemize}[leftmargin=1.8em, itemsep=2pt, topsep=2pt, parsep=0pt, partopsep=0pt]
\item \textbf{Theoretical Foundation.} We derive a fully closed-form training objective for \revr{an unsupervised} Student's $t$-mixture \revr{deep generative model (a heavy-tailed mixture VAE) and, to our knowledge, are the first to apply it to image prototype discovery}. We expose Gaussian–Gamma conjugacy through an auxiliary-variable construction. This formulation avoids biased variational approximations; furthermore, it ensures that the standard Gaussian-mixture case is recovered exactly as a limiting distribution when the degrees of freedom approach infinity.

\item \textbf{Inherent Noise Immunity.} We show that treating precision as a latent variable naturally yields an intrinsic reweighting mechanism. Prototype gradients are automatically downweighted based on a sample's distance from its cluster center, recovering classical robust M-estimators. This enables the model to learn adaptive, per-cluster degrees of freedom that handle varying noise levels without manual hyperparameter tuning.

\item \textbf{Empirical Performance.}The proposed model achieves higher \revr{prototype assignment accuracy and more distinctly separated latent representations}, improves training stability over state-of-the-art baselines on standard image benchmarks\revr{, matches the training cost of Gaussian VaDE or single-step generative models, and enables $33$--$36\times$ faster inference than a latent-diffusion baseline (e.g., with a step size exceeding 100)}. Learned prototypes preserve class-specific features under additive and structured noise, while baseline generative methods often collapse into blurred, non-discriminative templates.

\item \textbf{Clinical Utility.} In real-world clinical practice (e.g., during the diagnostic workup of neurodegenerative disorders), the model recovers anatomically faithful prototypes and well-calibrated uncertainty in cluster assignments. By preserving rare subtypes in high-noise regimes, it supports precise clinical analysis of disease heterogeneity where standard models might otherwise merge distinct atrophy patterns.
\end{itemize}

\section{Background: Variational Deep Embedding}\label{sec:background}
\paragraph{Mathematical Notation.}
We consider the task of embedding observations $\bx \in \R^D$ into a latent space $\bz \in \R^d$ characterized by \revr{mixture-component (prototype)} labels $c \in \{1,\dots,K\}$. We denote a Gaussian distribution as $\N(\bmu, \bm{\Lambda}^{-1})$ with mean $\bmu$ and precision $\bm{\Lambda}$. The Student distribution $\St_\nu(\bmu, \bm{\Lambda}^{-1})$ is defined by degrees of freedom $\nu$, location $\bmu$, and precision $\bm{\Lambda}$. We adopt the parameterization of the shape-rate for the Gamma distribution, where $u \sim \Gam(\alpha, \beta)$ denotes the density $p(u) \propto u^{\alpha-1} e^{-\beta u}$ with the mean $\alpha/\beta$. The digamma function is denoted by $\psi$.

\paragraph{The VaDE Framework.}
Variational Deep Embedding (VaDE) \citep{jiang2017variational} defines a generative process where cluster assignments and latent codes are sampled as follows:
\begin{equation}\label{eq:vade-gen}
c \sim \mathrm{Cat}(\bm{\pi}),\quad
\bz \mid c \sim \N(\bmu_c, \bm{\Lambda}_c^{-1}),\quad
\bx \mid \bz \sim p_\theta(\bx \mid \bz).
\end{equation}
\rev{Here, $\mathrm{Cat}(\bm{\pi})$ is the categorical distribution
over $K$ components, with mixing probabilities $\pi_c\geq0$ and $\sum_{c=1}^{K}\pi_c=1$.} The inference model is parameterized by a diagonal Gaussian encoder $q_\phi(\bz\mid\bx) = \N(\bmu_\phi, \mathrm{diag}(\bm{\sigma}^2_\phi))$. Cluster responsibilities $\gamma_c \coloneqq q(c \mid \bx)$ are computed \revr{via Bayes' rule}
\begin{equation}\label{eq:gamma-def}
\gamma_c = \frac{\pi_c\, p(\bz \mid c)}{\sum_{k=1}^K \pi_k\, p(\bz \mid k)},
\end{equation}
where $\bz$ is a sample from the encoder. The lower bound of the VaDE evidence (ELBO) is given by:
\begin{equation}\label{eq:vade-elbo}
\mathcal{L}_{\text{VaDE}}(\bx) = \E_{q_\phi}[\log p_\theta(\bx \mid \bz)] - \sum_{c=1}^K \gamma_c \left[ \KL[q_\phi(\bz \mid \bx) \,\|\, \N(\bmu_c, \bm{\Lambda}_c^{-1})] - \log \pi_c \right].
\end{equation}
This objective is computationally efficient because the KL divergence between two Gaussians admits a closed-form expression \cite{dilokthanakul2016deep,kingma2013auto}. 
\paragraph{The Analytic Obstacle for Robust Priors.}
Replacing the Gaussian prior in \eqref{eq:vade-gen} with a Student's $t$ prior $\St_{\nu_c}(\bmu_c, \bm{\Lambda}_c^{-1})$ introduces robustness to outliers due to the distribution's heavy tails~\citep{mclachlan2019finite}. However, this substitution renders the KL divergence in \eqref{eq:vade-elbo} intractable, as the integral $\int q_\phi(\bz\mid\bx) \log \St_{\nu_c}(\bz) d\bz$ has no closed-form solution. Existing methods often rely on numerical approximations or moment-matching surrogates \citep{kobayashi2019vade,kim2024t3vae}, which may introduce gradient variance or bias\revr{; a notable exception is the supervised text model of~\cite{boenninghoff2020variational}}. We present a methodology that retains a strictly closed-form ELBO under Student's $t$ priors.

\section{Methodology}

\label{sec:method}

\subsection{Generative Model}\label{subsec:model}
\begin{figure}[!t]
    \centering
    \includegraphics[width=.8\linewidth]{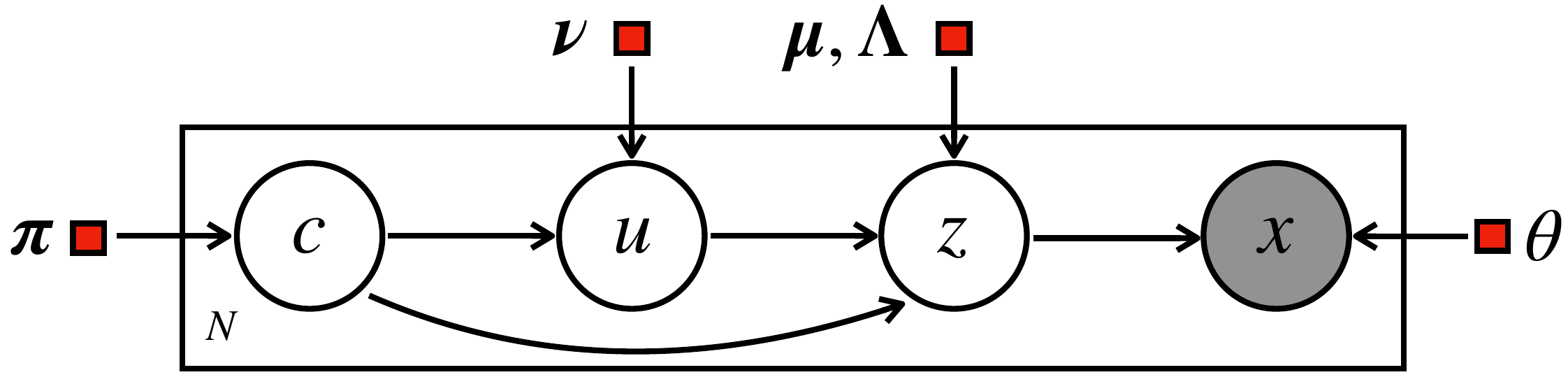}
\caption{Probabilistic graphical model of \textbf{Pro-\texorpdfstring{$t$}{t}-VaDE} as a \emph{scale mixture of Gaussians}. Shaded nodes are observed, white nodes latent, red squares global parameters, and plate $N$ indicates per-sample replication. Given cluster $c$, auxiliary precision $u\sim\Gam(\nu_c/2,\nu_c/2)$ scales $\bz$ covariance to $(u\bm{\Lambda}_c)^{-1}$: small $u$ models tails, large $u$ concentrates near the center. Marginalizing $u$ yields the Student's $t$ prior $\bz\mid c\sim\St_{\nu_c}(\bmu_c,\bm{\Lambda}_c^{-1})$. The encoder maps $\bx$ to $q_\phi(\bz\mid\bx)$, the mixture gives soft prototype assignments, and the shared decoder maps $\bz$ to reconstructions and $\bmu_c$ to image prototypes.}
    \label{fig:pgm}
\end{figure}

Our construction rests on a classical scale-mixture identity~\citep{andrews1974scale}, central to robust \rev{\revr{variational} Bayes (VB)} mixtures~\citep{svensen2005robust, peel2000robust}\revr{ and hierarchical regularization priors in medical image analysis~\citep{wang2021bayesian}}:
\begin{equation}\label{eq:smg-identity}
u \sim \Gam\!\left(\tfrac{\nu}{2},\tfrac{\nu}{2}\right),\quad
\bz \mid u \sim \N\!\bigl(\bmu, (u\bm{\Lambda})^{-1}\bigr)
\quad\Longrightarrow\quad
\bz \sim \St_\nu\!\bigl(\bmu, \bm{\Lambda}^{-1}\bigr).
\end{equation}
The auxiliary precision $u$ plays a dual role. Analytically, it decomposes the heavy-tailed prior into a Gaussian–Gamma hierarchy that admits closed-form inference. Statistically, it acts as a per-sample inlier weight: large $u$ tightens the conditional Gaussian around an inlier observation, while small $u$ inflates it to absorb tail observations without biasing the cluster mean. We adopt this identity in a per-cluster form to obtain the hierarchical generative model
\begin{equation}\label{eq:gen}
c \sim \mathrm{Cat}(\bm{\pi}),\quad
u \mid c \sim \Gam\!\left(\tfrac{\nu_c}{2},\tfrac{\nu_c}{2}\right),\quad
\bz \mid u, c \sim \N\!\bigl(\bmu_c,(u\bm{\Lambda}_c)^{-1}\bigr),\quad
\bx \mid \bz \sim p_\theta(\bx \mid \bz),
\end{equation}
illustrated in Fig.~\ref{fig:pgm}. Marginalizing $u$ recovers the per-cluster Student's $t$ prior $\bz \mid c \sim \St_{\nu_c}(\bmu_c, \bm{\Lambda}_c^{-1})$, so \eqref{eq:gen} is statistically equivalent to a $t$-mixture VaDE while preserving conditionally Gaussian latents. The degrees of freedom $\nu_c > 0$ are learned per cluster, so each prototype adapts its tail behavior to the empirical noise level of its cluster.

\subsection{Closed-Form ELBO via Analytical Collapsing}\label{subsec:elbo}

Substituting the Student-$t$ prior of \eqref{eq:gen} into the variational objective introduces a fundamental obstacle: the marginal divergence $\KL\!\bigl[q_\phi(\bz\mid\bx)\,\|\,\St_{\nu_c}(\bmu_c,\bm{\Lambda}_c^{-1})\bigr]$ admits no closed form since it requires evaluating $\E_{q_\phi}\!\bigl[\log(1 + \bar\Delta_c^2/\nu_c)\bigr]$. We resolve this \emph{variational intractability} through the \emph{analytical collapsing} of the auxiliary variable $u$, exploiting Gaussian--Gamma conjugacy to recover a fully analytic ELBO with exact gradients.

We employ the structured variational posterior
\begin{equation}\label{eq:variational}
q(\bz, u, c \mid \bx) \,=\, q_\phi(\bz\mid\bx)\, q(u\mid\bx,c)\, q(c\mid\bx),
\end{equation}
in which $q_\phi(\bz\mid\bx) = \N\!\bigl(\bmu_\phi(\bx), \mathrm{diag}(\bm{\sigma}_\phi^2(\bx))\bigr)$ is the amortized encoder and $\gamma_c \equiv q(c\mid\bx)$ is the cluster posterior. Defining the expected squared Mahalanobis distance under $q_\phi$,
\begin{equation}\label{eq:Delta}
\bar\Delta_c^2 \,\equiv\, \sum_{j=1}^{d} \Lambda_{c,j}\!\left[\sigma_{\phi,j}^2 + (\mu_{\phi,j} - \mu_{c,j})^2\right],
\end{equation}
Gaussian--Gamma conjugacy yields the optimal posterior in closed form,
\begin{equation}\label{eq:q-u-star}
q^\star(u\mid\bx,c) = \Gam(\tilde\alpha_c, \tilde\beta_c),
\qquad
\tilde\alpha_c = \tfrac{\nu_c + d}{2},\qquad
\tilde\beta_c  = \tfrac{\nu_c + \bar\Delta_c^2}{2},
\end{equation}
with sufficient statistics
\begin{equation}\label{eq:u-moments}
\E[u] \,=\, \frac{\nu_c + d}{\nu_c + \bar\Delta_c^2},
\qquad
\E[\log u] \,=\, \psi(\tilde\alpha_c) - \log\tilde\beta_c.
\end{equation}

Substituting $q^\star$ into the joint variational objective and integrating $u$ analytically, we obtain the central result of this section: a fully closed-form ELBO for the Student's $t$-mixture model,
\begin{equation}\label{eq:our-elbo}
\mathcal{L}(\bx) \,=\, \E_{q_\phi}\!\bigl[\log p_\theta(\bx\mid\bz)\bigr]
\,-\, \sum_{c=1}^{K} \gamma_c\!\left[\widetilde{\KL}_{\mathrm{Gauss}}^{(c)} + \KL_{\mathrm{Gamma}}^{(c)}\right]
\,-\, \KL\!\bigl[q(c\mid\bx)\,\|\,p(c)\bigr],
\end{equation}
All terms are evaluated analytically, with no sampling of $u$ or KL approximations. Because the KL for the marginal Student-$t$ distribution $\bz \mid c$ is intractable, it must be split into two KL terms. Introducing an auxiliary variable yields a tractable pair $(\widetilde{\KL}_{\mathrm{Gauss}}, \KL_{\mathrm{Gamma}})$ whose expected sum upper-bounds the marginal KL and provides exact gradients. The three terms of \eqref{eq:our-elbo} are
\begin{align}
&\widetilde{\KL}_{\mathrm{Gauss}}^{(c)} 
= \tfrac{1}{2}\!\left[\E[u]\,\bar\Delta_c^2 \,-\, d\,\E[\log u] \,-\, \textstyle\sum_j\log\Lambda_{c,j} \,-\, \sum_j\log\sigma_{\phi,j}^2 \,-\, d\right], \label{eq:kl-gauss}\\[2pt]
&\KL_{\mathrm{Gamma}}^{(c)} 
= \tfrac{d}{2}\psi(\tilde\alpha_c) \,+\, \log\tfrac{\Gamma(\nu_c/2)}{\Gamma(\tilde\alpha_c)} \,+\, \tfrac{\nu_c}{2}\log\!\left(1 + \tfrac{\bar\Delta_c^2}{\nu_c}\right) \,-\, \tfrac{1}{2}\E[u]\,\bar\Delta_c^2, \label{eq:kl-gamma}\\[2pt]
&\KL\!\bigl[q(c\mid\bx)\,\|\,p(c)\bigr]
= \textstyle\sum_{c=1}^{K} \gamma_c\bigl(\log\gamma_c - \log\pi_c\bigr). \label{eq:kl-cat}
\end{align}
Eq.~\eqref{eq:kl-gauss} is a Gaussian KL where the prior precision is scaled by the inferred inlier weight $\E[u]$: distant samples have smaller $\E[u]$ and thus a weaker influence on the cluster mean, matching the Iteratively Reweighted Least Squares (IRLS) dynamics in \eqref{eq:grad-mu}. Eq.~\eqref{eq:kl-gamma} adjusts the per-cluster degrees of freedom via $\tfrac{\nu_c}{2}\log\!\left(1 + \bar\Delta_c^2/\nu_c\right)$, the Student's $t$ tail penalty, so $\nu_c$ can be learned by gradient descent instead of being fixed. Eq.~\eqref{eq:kl-cat} is the standard categorical divergence between the cluster posterior and prior. Full derivations, expectation identities, and the extension to full-covariance $\bm{\Lambda}_c$ appear in the appendix.

\rev{\paragraph{Computational Cost.}
\revr{Notably}, The auxiliary scale variable needs no extra networks or Monte Carlo estimates, since $\E[u]$ and $\E[\log u]$ have closed forms. Compared with Gaussian VaDE, it adds only $K$ trainable scalars. Overhead scales as $O(BK)$, while diagonal mixture evaluation remains $O(BKd)$ for batch size $B$, $K$ components, and latent dimension $d$. \revr{Thus, it improves robustness to heavy-tailed noise with essentially the same asymptotic complexity as a Gaussian mixture prior.}}

\paragraph{Intrinsic Robustness and Gradient Dynamics.}\label{subsec:irls}
The model’s robustness follows from the gradient of \eqref{eq:our-elbo} w.r.t. $\bmu_c$:
\begin{equation}\label{eq:grad-mu}
\nabla_{\bmu_c}\mathcal{L} \propto \gamma_c \, \E[u] \, \bm{\Lambda}_c (\bmu_\phi - \bmu_c), \quad \text{where} \quad \E[u] = \frac{\nu_c + d}{\nu_c + \bar\Delta_c^2}.
\end{equation}
This gives an IRLS scheme \citep{lange1989robust}. For inliers with $\bar\Delta_c^2 \approx d$, $\E[u]\approx 1$, recovering the Gaussian VaDE update. For outliers with $\bar\Delta_c^2 \gg \nu_c$, $\E[u]\to 0$, so they have negligible effect on the cluster prototype.\footnote{This matches classical IRLS for robust $t$-mixture estimation, where $w_i \propto (\nu+d)/(\nu+\Delta_i^2)$ and prototypes are updated by IRLS \citep{lange1989robust, peel2000robust,ba2013convergence}, yielding vanishing outlier influence.} Unlike standard IRLS, these weights come from the variational objective, not an imposed rule.

\paragraph{Plug-and-Play Generality.}\label{subsec:arch}
\rev{The conjugate construction can be incorporated into encoder-decoder
architectures satisfying the stated latent-distribution assumptions. Our experiments use ResNet-based backbones, and the evaluation of method with additional architectures is future work. A high-level forward pass is given in Alg.~\ref{alg:fwd}.} \revr{In this sense, the Student's $t$ mixture prior acts as a drop-in, closed-form replacement for the Gaussian mixture prior of VaDE-style generative models.}

\begin{algorithm}[H]
\caption{Forward pass of our model.}\label{alg:fwd}
\begin{algorithmic}[1]
\REQUIRE Input $\bx$; encoder $q_\phi$; decoder $p_\theta$; \revr{mixture} parameters $\{\bmu_c,\bm{\Lambda}_c,\pi_c,\nu_c\}_{c=1}^K$.
\STATE $(\bmu_\phi,\bm{\sigma}^2_\phi)\leftarrow\mathrm{Enc}(\bx)$;\quad $\bz\leftarrow\bmu_\phi+\bm{\sigma}_\phi\odot\bm{\epsilon}$,\ $\bm{\epsilon}\sim\N(\bm{0},\bm{I})$.
\STATE $\hat{\bx}\leftarrow\mathrm{Dec}(\bz)$;\quad $\mathcal{L}_{\mathrm{rec}}\leftarrow-\log p_\theta(\bx\mid\bz)$.
\STATE $\gamma_c \propto \pi_c\,\St_{\nu_c}(\bz \mid \bmu_c, \bm{\Lambda}_c^{-1})$\ for all $c$.
\FOR{$c=1,\dots,K$}
    \STATE Compute $\bar\Delta_c^2$ via \eqref{eq:Delta}; $\E[u], \E[\log u]$ via \eqref{eq:u-moments}.
    \STATE Evaluate $\widetilde{\KL}^{(c)}_{\mathrm{Gauss}}$ \eqref{eq:kl-gauss} and $\KL^{(c)}_{\mathrm{Gamma}}$ \eqref{eq:kl-gamma}.
\ENDFOR
\STATE Compute $\KL_{\mathrm{Cat}} \leftarrow \sum_c \gamma_c\bigl(\log\gamma_c - \log\pi_c\bigr)$ via \eqref{eq:kl-cat}.
\STATE \textbf{return}\ $\mathcal{L} = \mathcal{L}_{\mathrm{rec}} + \sum_c \gamma_c\!\left[\widetilde{\KL}^{(c)}_{\mathrm{Gauss}} + \KL^{(c)}_{\mathrm{Gamma}}\right] + \KL_{\mathrm{Cat}}$.
\end{algorithmic}
\end{algorithm}

\section{Experiments}\label{sec:exp}
We evaluate our method on three axes: \revr{prototype assignment accuracy} under noise (\S\ref{subsec:exp-clustering}), prototype quality \revr{and representation separability} (\S\ref{subsec:exp-proto}), and clinical validity in a real-world neurodegenerative cohort (\S\ref{subsec:exp-clinical}). We describe the setup, datasets, baselines, and evaluation protocols once below; later sections report only results.

\subsection{Setup}\label{subsec:exp-setup}

We evaluate on two \revr{standard} image benchmarks and one real-world medical imaging cohort. MNIST~\citep{lecun1998gradient} and Fashion-MNIST~\citep{xiao2017fashion} both contain $60{,}000$ training and $10{,}000$ test grayscale images of size $28\!\times\!28$ across $K\!=\!10$ classes; together they probe \revr{prototype discovery} data of qualitatively different complexity (handwritten digits vs. clothing categories). OASIS Brain MRI~\citep{lamontagne2019oasis} provides $T_1$-weighted brain volumes of size $128^3$ from a subset of $253$ subjects aged $60$--$95$, including healthy controls and Alzheimer's disease patients, used to assess the model's ability to recover anatomically faithful disease prototypes under realistic acquisition noise.

\paragraph{Baselines.} We benchmark against six representative deep clustering \revr{and generative mixture} models spanning key architectural and methodological choices. VaDE~\cite{jiang2017variational} is a Gaussian-prior variational deep clustering model. Contra-VaDE \cite{li2021contrastive} adds a contrastive objective to separate latent clusters.\revr{ }LDDPM-VaDE \cite{yan2023clusterddpm} adapts ClusterDDPM by running diffusion in latent space. The other three baselines use a Student’s $t$ prior, addressing variational intractability differently: $t$-VaDE-log uses Monte Carlo log-gradient estimators for $\E_{q_\phi}[\log(1+\bar\Delta^2/\nu)]$ \cite{kim2024t3vae}; $t$-VaDE-EM alternates variational updates with EM cluster-parameter updates \cite{kobayashi2019vade}; and $t$-VaDE-2Ph trains in two phases, transitioning from a Gaussian to a Student’s $t$ prior. Our method is the only fully end-to-end approach with a closed-form ELBO under a Student’s $t$ prior.

\paragraph{Implementation.} All baselines and our model use the same ResNet-based encoder–decoder backbone with a symmetric decoder. We train for $100$ epochs using AdamW with a learning rate of $10^{-3}$ for all parameters, weight decay $10^{-4}$, and batch size $32$. The \revr{number of prototypes} is fixed to the ground-truth $K\!=\!10$ on MNIST and Fashion-MNIST and $K=2$ for the
 brain MRI dataset. In our model, $\nu_c$ is softplus-parameterized and initialized at $\nu_c\!=\!5$ for every component. \rev{Sensitivity to the initialization of $\nu_c$ can be \revr{found in the} appendix}. All experiments run on 4\,$\times$\,NVIDIA 5880 Ada GPUs with 48\,GB memory each.

\paragraph{Evaluation protocols.} All experiments follow the same protocol. For \emph{\revr{prototype assignment accuracy}} (\S\ref{subsec:exp-clustering}), we corrupt MNIST and Fashion-MNIST with Cauchy noise at scales $\gamma\!\in\!\{0.02,0.04,\ldots,0.14\}$ and report \revr{standard clustering metrics} ACC, ARI, and NMI (higher is better). Cauchy noise is a strict stress test because its undefined moments violate Gaussian-prior assumptions of finite second-order statistics. For \emph{prototype quality} (\S\ref{subsec:exp-proto}), under the same noise we measure: (i) PSNR between matched prototypes learned from noisy vs.\ clean data; (ii) latent separability using Silhouette (Sil, $\uparrow$), Davies--Bouldin (DB, $\downarrow$), and $t$-SNE; and (iii) training stability via the rate of \emph{cluster collapse} (two or more prototypes converging to the same latent location) over $118$ runs. For \emph{clinical validity} (\S\ref{subsec:exp-clinical}), we apply nine realistic Brain MRI corruptions (Gaussian, Rician, and bias-field; mild/medium/severe) and evaluate the same metrics.

\subsection{\texorpdfstring{\revr{Prototype Assignment Accuracy Under Noise}}{Prototype Assignment Accuracy Under Noise}}\label{subsec:exp-clustering}

\begin{figure}[!htb]
    \centering
    \includegraphics[width=\linewidth]{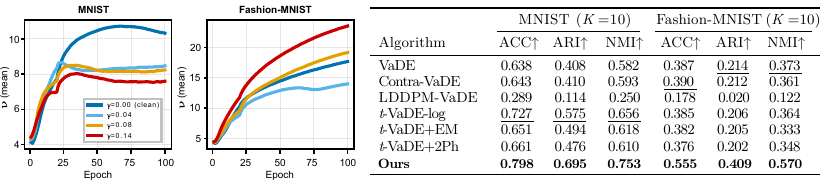}
    \caption{\textbf{Left:} Mean of the learned per-cluster degrees of freedom $\nu_c$ across training epochs on MNIST and Fashion-MNIST under four Cauchy-noise levels $\gamma \in \{0.00, 0.04, 0.08, 0.14\}$; all runs initialize $\nu_c = 5$. \textbf{Right:} \revr{Prototype assignment accuracy} (ACC, ARI, NMI) averaged over noise levels $\gamma \in \{0.02, \ldots, 0.14\}$ on MNIST and Fashion-MNIST. Best in \textbf{bold}, second \underline{underlined}.}
    \label{fig:nu_para}
\end{figure}

Fig.~\ref{fig:nu_para}~(left) tracks $\nu_c$ during training. On MNIST, $\nu_c$ converges near $10$ on clean data and decreases to about $4$ at $\gamma=0.14$, consistent with robust $t$-estimation~\citep{lange1989robust, peel2000robust}: heavier tails call for smaller $\nu_c$ to absorb outliers without biasing the prototypes. Fashion-MNIST shows the opposite trend, with $\nu_c$ rising as noise grows. This reflects its cluster geometry: classes like \emph{pullover}, \emph{shirt}, and \emph{coat} already exhibit large intra-class variation, so $\nu_c$ stays small even on clean inputs; adding noise triggers the IRLS reweighting in~\eqref{eq:grad-mu} to downweight corrupted samples, tightening the remaining geometry and pushing $\nu_c$ toward the Gaussian regime.\revr{ }These results show $\nu_c$ adapts to dataset geometry, a capability absent in fixed-$\nu$ formulations.

Fig.~\ref{fig:nu_para}~(right) reports \revr{prototype assignment accuracy} averaged over Cauchy-noise levels $\gamma\in\{0.02,\ldots,0.14\}$. Our method leads on every metric across both datasets. Variants with fixed or externally scheduled $\nu$ (t-VaDE-log, t-VaDE+EM, t-VaDE+2Ph) improve over VaDE on MNIST but stall on the more heterogeneous Fashion-MNIST, while LDDPM-VaDE collapses under noise on both. The gap is largest on Fashion-MNIST, where per-cluster adaptation of $\nu_c$ delivers a $\sim$$42\%$ relative ACC gain over the strongest baseline, confirming that jointly learning $\nu_c$, rather than tuning a global heaviness parameter, is what drives robustness across heterogeneous geometries. \rev{Over five random seeds, our method achieves statistically significant gains over the strongest baselines: +0.079 on MNIST (95\% CI [0.052, 0.106], p=0.0025) and +0.116 on Fashion-MNIST (95\% CI [0.032, 0.200], p=0.018). See Appendix for the five-seed results and paired tests with Holm correction.}  

\begin{figure}[!htb]
    \centering
    \includegraphics[width=1.0\linewidth]{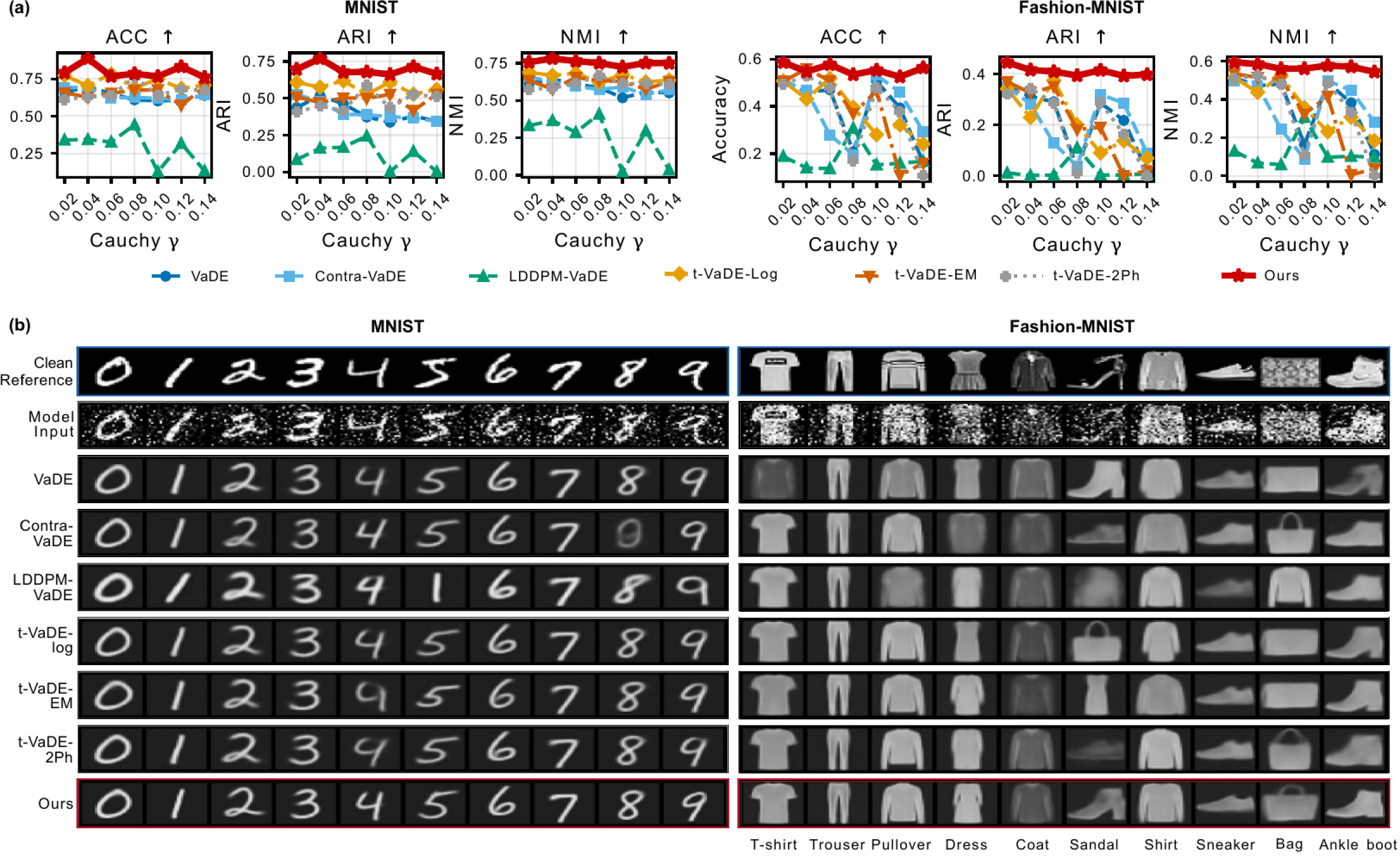}
    \caption{\rev{Top: quantitative evaluation of all models under increasing noise corruption of images. Bottom: prototype visualization across all models when image is under severe corruption $\gamma = 0.12$. Our approach preserves crisp, distinguishable prototypes without duplication.}}
    \label{fig:mnist-fmnist}
\end{figure}
Fig.~\ref{fig:mnist-fmnist} (top) reports ACC, ARI, and NMI as a function of noise level on both benchmarks; \revr{across} all noise scales, our method (red) maintains a near-flat performance curve, while every baseline degrades sharply and most fail for $\gamma\!\geq\!0.08$. Averaged over the seven noise levels, our model attains ACC of $0.798$ on MNIST and $0.555$ on Fashion-MNIST against best-baseline values of $0.727$ and $0.390$ respectively, with similarly large margins on ARI and NMI. 

\rev{\paragraph{Runtime Efficiency.}
The auxiliary-variable construction has a modest measured cost: training takes 33.21/32.00 ms per update on MNIST/Fashion-MNIST, compared with 27.74/27.05 for VaDE and 31.99/30.20 for LDDPM-VaDE. In a separate benchmark executing the complete 100-step reverse trajectory of the same latent-diffusion network, end-to-end assignment takes 0.142/0.142 ms per sample for LDDPM-VaDE and 0.004/0.004 for ours, corresponding to $33\times$/$36\times$ speedups. These speedups concern the specified reverse-inference path, not training updates. Protocols and training costs are reported in the appendix.} \revr{Together with the accuracy gains above, this shows that our generative framework attains heavy-tailed robustness without sacrificing efficiency.}

\subsection{\texorpdfstring{Prototype Quality \revr{and Representation Separability}}{Prototype Quality and Representation Separability}}\label{subsec:exp-proto}

\begin{figure}[!htb]
    \centering
    \includegraphics[width=0.90\linewidth]{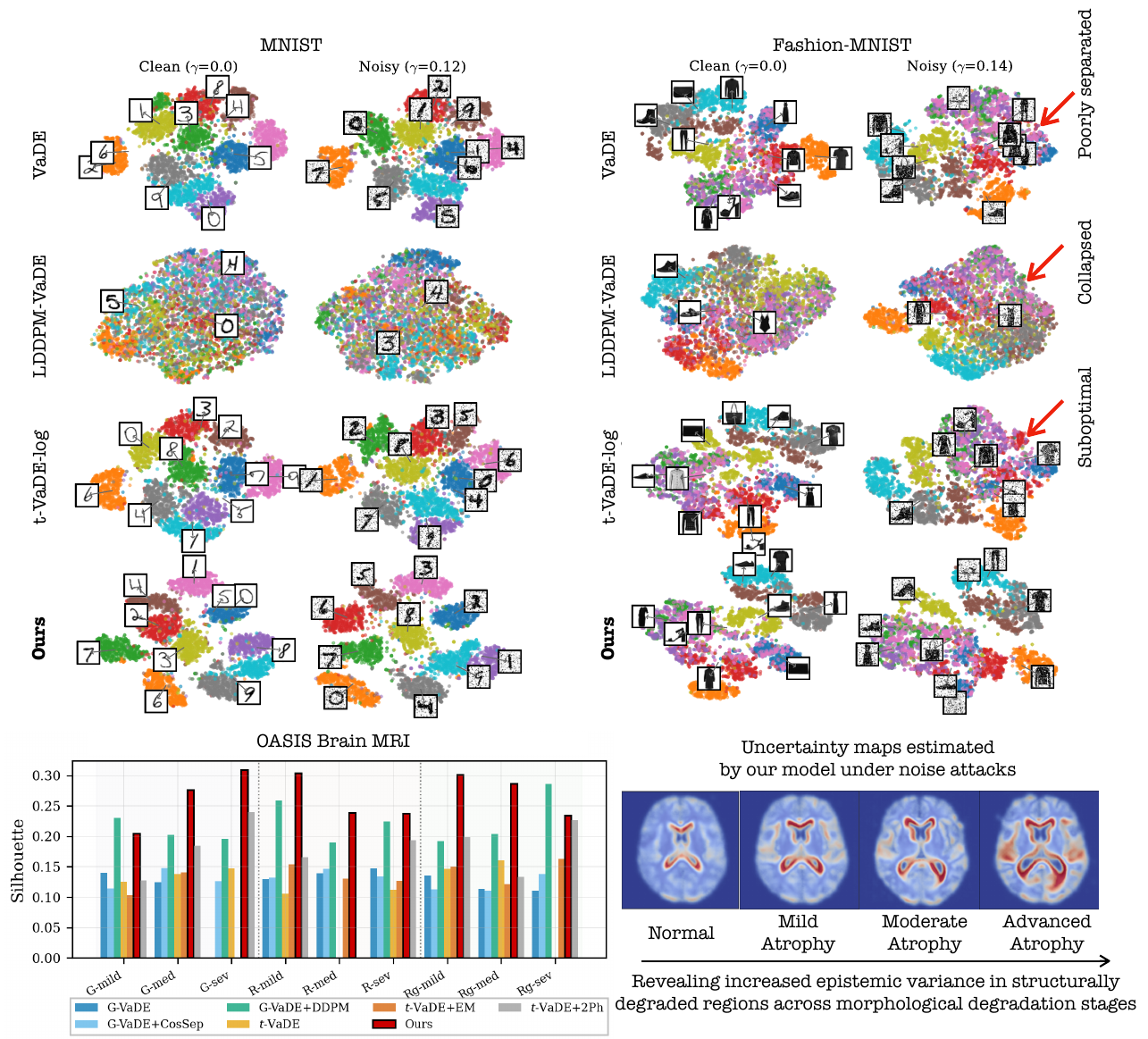}
    \caption{Top: $t$-SNE of latent representations from VaDE, LDDPM-VaDE, t-VaDE-log, and ours on MNIST and Fashion-MNIST with clean ($\gamma=0.0$) and noisy inputs ($\gamma=0.12$ MNIST, $0.14$ Fashion-MNIST). Bottom-left: Silhouette on OASIS Brain MRI under nine realistic corruptions (Gaussian, Rician, bias-field; mild/medium/severe). Bottom-right: per-voxel uncertainty maps from our model across \rev{four groups of anatomy-associated patterns.}}
    \label{fig:tsne-collapse}
\end{figure}
\rev{Fig.~\ref{fig:mnist-fmnist} (bottom) shows decoded prototypes at $\gamma=0.12$. Our method preserves recognizable digits on MNIST and distinguishable garments on Fashion-MNIST, while baseline prototypes are blurred or exhibit cluster collapse, with multiple prototypes converging to the same shape and leaving some classes unrepresented.

Fig.~\ref{fig:tsne-collapse} (top) shows t-SNE projections. On clean inputs, \revr{competing} models generally recover cluster structure. However, under noise, VaDE leaves clusters overlapping, LDDPM-VaDE collapses into an unstructured blob, and t-VaDE-log retains only blurred boundaries, while our model preserves well-separated clusters. The per-cluster $\nu_c$ downweights corrupted samples, anchoring both prototypes and latent geometry on clean class structure rather than noise.} \revr{The separated latent representations are thus a direct by-product of the heavy-tailed generative prior and underpin the quality of the discovered prototypes.}

\subsection{\rev{Anatomical Robustness and Medical Image Validation}}\label{subsec:exp-clinical}
\begin{figure}[!htb]
    \centering
    \includegraphics[width=.95\linewidth]{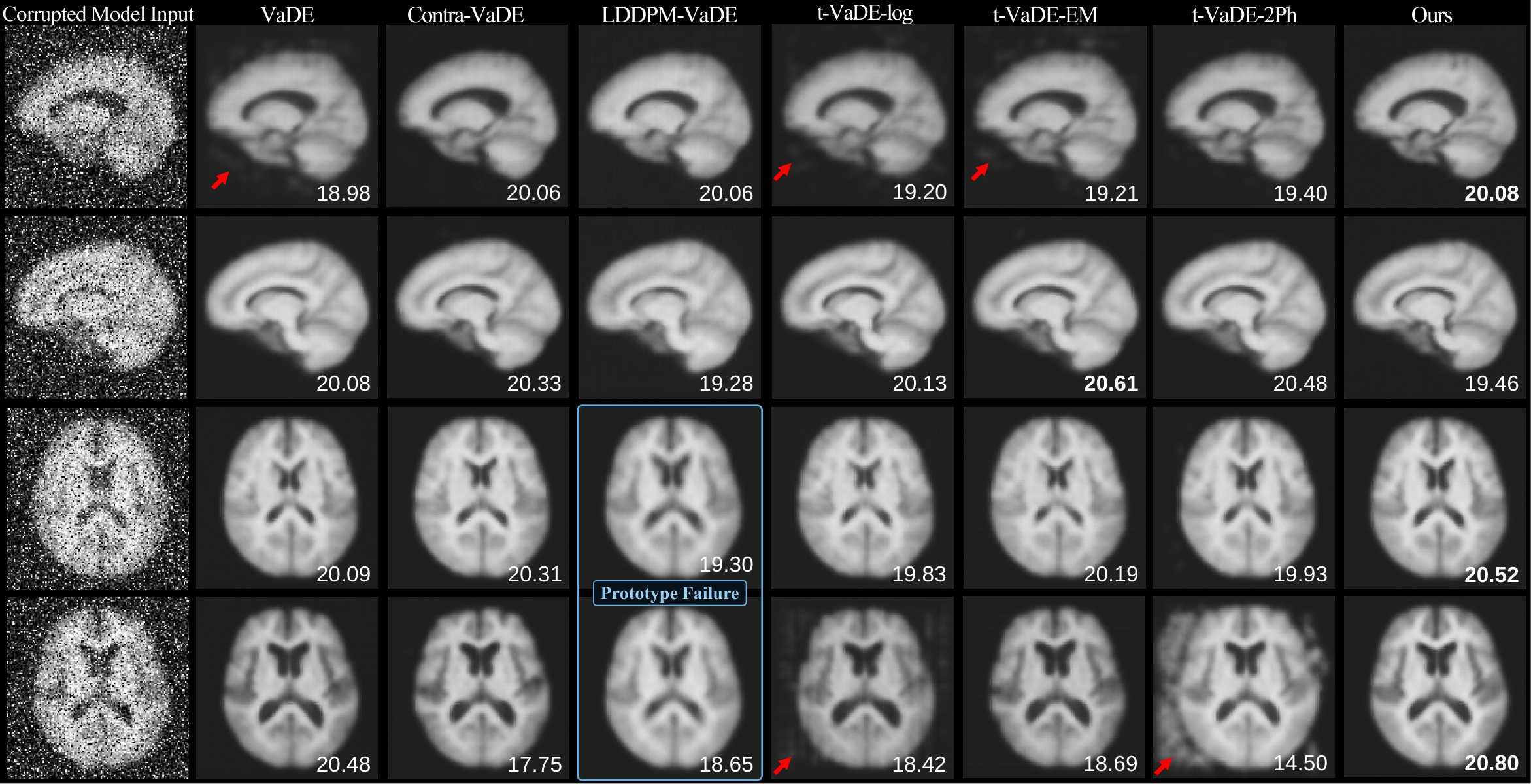}
    \caption{\rev{Visual comparison of learned atlases on OASIS dataset under Cauchy noise ($\gamma=0.12$). Col. 1 shows corrupted inputs; cols. 2–8 show cluster-mean atlases reconstructed by different methods. Rows 1–2 are sagittal atlases for two clusters; rows 3–4 are coronal atlases. White numbers (bottom right) report per-cluster PSNR (dB) vs. each method’s clean atlas after one-to-one cluster matching (higher is better); best per row is bold. Red arrows mark baseline artifacts. The blue box shows an LDDPM-VaDE failure: the two coronal cluster atlases are nearly identical, indicating prototype collapse. Aggregate PSNR and Silhouette results are reported in Tab.~\ref{tab:main}(a)}.}
    \label{fig:atlas}
\end{figure}

Fig.~\ref{fig:atlas} shows prototypes learned from heavily noise-corrupted OASIS Brain MRI (left). Our prototypes are clean and artifact-free, with sharp sagittal and coronal boundaries. VaDE, Contra-VaDE, and LDDPM-VaDE yield blurred prototypes with smeared cortex and washed-out ventricles; $t$-VaDE variants recover more anatomy but retain noise and cortical ghosting. Downweighting corrupted voxels via per-cluster $\nu_c$ anchors prototypes to clean anatomy rather than noise patterns, \rev{which may benefit downstream tasks such as atlas-based registration and population-level subtype analysis.}

Fig.~\ref{fig:tsne-collapse} (bottom-left) shows Silhouette scores on OASIS Brain MRI across nine realistic corruptions (Gaussian, Rician, and bias-field noise) at mild, medium, and severe levels. Our method attains the highest Silhouette in all conditions, with the largest gains under severe noise (G-sev, R-sev, Rg-sev). Robustness on synthetic Cauchy noise thus extends to clinically realistic artifacts, where labels are unavailable and latent separability is the main evaluation criterion.

\rev{Fig.~\ref{fig:tsne-collapse}~(bottom-right) shows per-pixel standard deviation of four groups of anatomy-associated patterns, each produced by averaging 32 decoded samples. From left to right, the estimated variance grows in atrophied regions such as the ventricles and surrounding cortical tissue. The heavy-tailed $t$-distribution thus yields not only robustness but also may potentially benefit clinically interpretable epistemic uncertainty, flagging regions where pathological change reduces the model's confidence.}

\begin{table}[!htb]
\centering
\caption{\revr{Prototype separability and reconstruction quality} on OASIS at $K=2$, averaged over sagittal/axial/coronal views. \textbf{(a)} Sil, DB, and PSNR at four noise levels; PSNR is reported only for corrupted conditions relative to matched clean-run prototypes (clean shown as \texttt{--}). Collapsed runs (single cluster; Sil undefined) are excluded; cells with any excluded run are \textit{italic}$^{\dagger}$. \textbf{(b)} Collapse count and rate per method over 118 runs. Best in \textbf{bold}.}
\label{tab:main}
\captionsetup[subtable]{font=footnotesize, skip=4pt, position=top, justification=centering}
\subcaptionbox{\revr{Prototype separability} and reconstruction at four noise levels.\label{tab:gamma_subset}}[0.78\linewidth]{}%
\hfill
\subcaptionbox{Model collapse over 118 runs.\label{tab:collapse}}[0.20\linewidth]{}
\par\smallskip
\scriptsize
\setlength{\tabcolsep}{1.5pt}
\begin{tabular*}{\linewidth}{@{\extracolsep{\fill}}l ccc ccc ccc ccc@{\hspace{14pt}}cc@{}}
\cmidrule[\heavyrulewidth](r{7pt}){1-13}\cmidrule[\heavyrulewidth](l{0pt}){14-15}
\multirow{2}{*}{Method}
 & \multicolumn{3}{c}{$\gamma=0$ (clean)}
 & \multicolumn{3}{c}{$\gamma=0.02$}
 & \multicolumn{3}{c}{$\gamma=0.08$}
 & \multicolumn{3}{c}{$\gamma=0.12$}
 & \multirow{2}{*}{Count$\downarrow$} & \multirow{2}{*}{Rate (\%)$\downarrow$} \\
\cmidrule(lr){2-4} \cmidrule(lr){5-7} \cmidrule(lr){8-10} \cmidrule(lr){11-13}
 & Sil$\uparrow$ & DB$\downarrow$ & PSNR$\uparrow$
 & Sil$\uparrow$ & DB$\downarrow$ & PSNR$\uparrow$
 & Sil$\uparrow$ & DB$\downarrow$ & PSNR$\uparrow$
 & Sil$\uparrow$ & DB$\downarrow$ & PSNR$\uparrow$ & & \\
\cmidrule[\lightrulewidth](r{7pt}){1-13}\cmidrule[\lightrulewidth](l{0pt}){14-15}
VaDE            & 0.170 & 1.913 & --   & 0.165 & 1.952 & 25.75          & 0.181 & 1.934 & 21.81          & 0.200 & 1.811 & 19.82 & 2 & 1.7 \\
Contra-VaDE     & 0.172 & 1.932 & --   & 0.168 & 1.966 & 22.02          & 0.176 & 2.039 & 20.45          & 0.215 & 1.759 & 19.57 & 1 & 0.8 \\
LDDPM-VaDE      & 0.238 & 1.618 & --   & 0.249 & 1.534 & \textbf{28.28} & 0.262 & 1.446 & 21.71          & \textit{0.309}$^{\dagger}$ & \textit{1.300}$^{\dagger}$ & \textit{19.74}$^{\dagger}$ & 16 & 13.6 \\
$t$-VaDE-log    & 0.165 & 1.979 & --   & 0.164 & 2.013 & 25.38          & 0.183 & 1.965 & 21.88          & 0.176 & 1.923 & 19.47 & 2 & 1.7 \\
$t$-VaDE+EM     & 0.167 & 2.055 & --   & 0.182 & 1.916 & 24.80          & 0.176 & 1.955 & 21.32          & 0.206 & 1.874 & 19.61 & 4 & 3.4 \\
$t$-VaDE+2Ph    & 0.248 & 1.520 & --   & 0.258 & 1.493 & 25.47          & 0.267 & 1.443 & \textbf{22.12} & 0.242 & 1.608 & 18.93 & 3 & 2.5 \\
\textbf{Ours}   & \textbf{0.326} & \textbf{1.181} & -- & \textbf{0.321} & \textbf{1.157} & 26.15 & \textbf{0.328} & \textbf{1.185} & 22.05 & \textbf{0.351} & \textbf{1.099} & \textbf{20.01} & \textbf{0} & \textbf{0.0} \\
\cmidrule[\heavyrulewidth](r{7pt}){1-13}\cmidrule[\heavyrulewidth](l{0pt}){14-15}
\end{tabular*}
\end{table}
Tab.~\ref{tab:main}(a) reports \revr{prototype separability and reconstruction quality} on OASIS at $K=2$, with metrics marginalized over the principal anatomical orientations to ensure the representativeness of the global volumetric structure. Our method consistently attains the best Sil and DB across all noise regimes, with both metrics remaining stable under increasing corruption $\gamma$ (Sil: $0.321$--$0.351$; DB: $1.099$--$1.185$), while baselines degrade substantially. On reconstruction fidelity, our approach yields the highest PSNR ($20.01$~dB) at ($\gamma=0.12$). The elevated Sil and DB of LDDPM-VaDE at $\gamma=0.12$ are artifacts of selective aggregation (marked $^{\dagger}$): that baseline frequently suffered mode collapse, and the reported values are computed only over its surviving runs.

Tab.~\ref{tab:main}(b) reports cluster-collapse rates across all $118$ trials. LDDPM-VaDE collapses on $13.6\%$ of runs, the remaining baselines on $0.8\%$--$3.4\%$, and our method on none. Combined with the stable Sil and DB above, this indicates that the per-cluster $\nu_c$ provides a reliability margin invisible to accuracy metrics alone. In a clinical setting, where each subject is a real patient and a degenerate \revr{solution} must be discarded rather than retried, a zero-collapse rate is the property that matters most.

\rev{\section{Discussion \& Conclusion}
\label{sec:conclusion}
In this paper, we introduce Pro-$t$-VaDE and show that heavy-tailed robustness and analytical tractability can be jointly achieved in deep generative \revr{mixture models for image prototype discovery}.By leveraging the Gaussian-Gamma conjugate structure of the Student's $t$-distribution, we derive a fully closed-form ELBO, which effectively eliminates the stochastic variance inherent in traditional gradient-based variational inference. Our framework establishes a rigorous link between deep latent variable models and classical M-estimation, providing a principled mechanism to navigate the trade-off between model expressivity and robustness. \revr{Across natural-image benchmarks and brain MRI, this yields accurate, efficient, and well-separated representations with a negligible overhead over a Gaussian mixture prior.} Empirical evaluations on neurodegenerative disease datasets show that our method preserves fine-grained anatomy-associated patterns that are often obscured by Gaussian-based models, offering a more granular lens into disease heterogeneity. It is worth noting that several aspects warrant future exploration. \revr{Our} method is particularly suited for scenarios where data contain \revr{heavy-tailed} noise yet analytical tractability is desired, such as clinical MRI with acquisition artifacts, where the per-cluster $\nu_c$ automatically downweights outliers. However, the current construction assumes a single elliptical component per cluster, and the \revr{number of prototypes} $K$ is fixed a priori rather than inferred via Bayesian nonparametric priors. Future efforts will extend this conjugate construction to broader scale mixtures\revr{, other latent-variable generative models,} and explore automatic model selection for dynamic \revr{prototype} discovery in longitudinal image analysis.
}

\begin{ack}
This work was supported by the National Natural Science Foundation of China (Serial No. 62406182), and Fundamental Research Program of Shanxi Province(Serial No. 202403021212176).
\end{ack}

\bibliographystyle{abbrv}
\bibliography{ref.bib}

@inproceedings{jiang2017variational,
  title     = {Variational Deep Embedding: An Unsupervised and Generative
               Approach to Clustering},
  author    = {Jiang, Zhuxi and Zheng, Yin and Tan, Huachun and Tang,
               Bangsheng and Zhou, Hanning},
  booktitle = {International Joint Conference on Artificial Intelligence
               (IJCAI)},
  year      = {2017}
}

@article{dilokthanakul2016deep,
  title   = {Deep Unsupervised Clustering with {G}aussian Mixture Variational
             Autoencoders},
  author  = {Dilokthanakul, Nat and Mediano, Pedro A. M. and Garnelo, Marta
             and Lee, Matthew C. H. and Salimbeni, Hugh and Arulkumaran,
             Kai and Shanahan, Murray},
  journal = {arXiv preprint arXiv:1611.02648},
  year    = {2016}
}

@article{andrews1974scale,
  title   = {Scale Mixtures of Normal Distributions},
  author  = {Andrews, David F. and Mallows, Colin L.},
  journal = {Journal of the Royal Statistical Society: Series B},
  volume  = {36},
  number  = {1},
  pages   = {99--102},
  year    = {1974}
}

@article{lange1989robust,
  title   = {Robust Statistical Modeling Using the {$t$} Distribution},
  author  = {Lange, Kenneth L. and Little, Roderick J. A. and Taylor,
             Jeremy M. G.},
  journal = {Journal of the American Statistical Association},
  volume  = {84},
  number  = {408},
  pages   = {881--896},
  year    = {1989}
}

@article{peel2000robust,
  title   = {Robust Mixture Modelling Using the {$t$} Distribution},
  author  = {Peel, David and McLachlan, Geoffrey J.},
  journal = {Statistics and Computing},
  volume  = {10},
  number  = {4},
  pages   = {339--348},
  year    = {2000}
}

@inproceedings{xie2016unsupervised,
  title     = {Unsupervised Deep Embedding for Clustering Analysis},
  author    = {Xie, Junyuan and Girshick, Ross and Farhadi, Ali},
  booktitle = {International Conference on Machine Learning (ICML)},
  year      = {2016}
}

@inproceedings{guo2017improved,
  title     = {Improved Deep Embedded Clustering with Local Structure
               Preservation},
  author    = {Guo, Xifeng and Gao, Long and Liu, Xinwang and Yin,
               Jianping},
  booktitle = {International Joint Conference on Artificial Intelligence
               (IJCAI)},
  year      = {2017}
}

@inproceedings{caron2018deep,
  title     = {Deep Clustering for Unsupervised Learning of Visual
               Features},
  author    = {Caron, Mathilde and Bojanowski, Piotr and Joulin, Armand
               and Douze, Matthijs},
  booktitle = {European Conference on Computer Vision (ECCV)},
  year      = {2018}
}

@inproceedings{caron2020unsupervised,
  title     = {Unsupervised Learning of Visual Features by Contrasting
               Cluster Assignments},
  author    = {Caron, Mathilde and Misra, Ishan and Mairal, Julien and
               Goyal, Priya and Bojanowski, Piotr and Joulin, Armand},
  booktitle = {Advances in Neural Information Processing Systems
               (NeurIPS)},
  year      = {2020}
}

@inproceedings{li2021contrastive,
  title     = {Contrastive Clustering},
  author    = {Li, Yunfan and Hu, Peng and Liu, Zitao and Peng, Dezhong
               and Zhou, Joey Tianyi and Peng, Xi},
  booktitle = {AAAI Conference on Artificial Intelligence (AAAI)},
  year      = {2021}
}

@inproceedings{caron2021emerging,
  title     = {Emerging Properties in Self-Supervised Vision Transformers},
  author    = {Caron, Mathilde and Touvron, Hugo and Misra, Ishan and
               J{\'e}gou, Herv{\'e} and Mairal, Julien and Bojanowski,
               Piotr and Joulin, Armand},
  booktitle = {IEEE International Conference on Computer Vision (ICCV)},
  year      = {2021}
}

@inproceedings{kim2024t3vae,
  title={{$t^3$}-Variational Autoencoder: Learning Heavy-tailed Data with Student's t and Power Divergence},
  author={Kim, Juno and Kwon, Jaehyuk and Cho, Mincheol and Lee, Hyunjong and Won, Joong-Ho},
  booktitle={International Conference on Learning Representations},
  volume={2024},
  pages={19580--19606},
  year={2024}
}

@inproceedings{kobayashi2019vade,
  title={Variational deep embedding with regularized student-t mixture model},
  author={Kobayashi, Taisuke},
  booktitle={International Conference on Artificial Neural Networks},
  pages={443--455},
  year={2019},
  organization={Springer}
}

@article{blaiotta2018generative,
  title   = {Generative Diffeomorphic Modelling of Large {MRI} Data Sets
             for Probabilistic Template Construction},
  author  = {Blaiotta, Claudia and Freund, Patrick and Cardoso,
             M. Jorge and Ashburner, John},
  journal = {NeuroImage},
  volume  = {166},
  pages   = {117--134},
  year    = {2018}
}

@article{rakic2025atlasmorph,
  title={AtlasMorph: Learning conditional deformable templates for brain MRI},
  author={Rakic, Marianne and Hoopes, Andrew and Abulnaga, Mazdak S and Sabuncu, Mert R and Guttag, John V and Dalca, Adrian V and Alzheimer’s Disease Neuroimaging Initiative and others},
  journal={Medical Image Analysis},
  volume={110},
  pages={103893},
  year={2026},
  publisher={Elsevier}
}

@inproceedings{song2024morphological,
  title={Morphological prototyping for unsupervised slide representation learning in computational pathology},
  author={Song, Andrew H and Chen, Richard J and Ding, Tong and Williamson, Drew FK and Jaume, Guillaume and Mahmood, Faisal},
  booktitle={Proceedings of the IEEE/CVF Conference on Computer Vision and Pattern Recognition},
  pages={11566--11578},
  year={2024}
}

@article{yan2023clusterddpm,
  title={ClusterDDPM: an EM clustering framework with denoising diffusion probabilistic models},
  author={Yan, Jie and Liu, Jing and Zhang, Zhong-yuan},
  journal={Information Sciences},
  pages={122518},
  year={2025},
  publisher={Elsevier}
}

@inproceedings{dasilva2024treediffusion,
  title={TreeDiffusion: Hierarchical Generative Clustering for Conditional Diffusion},
  author={da Silva Gon{\c{c}}alves, Jorge and Manduchi, Laura and Vandenhirtz, Moritz and Vogt, Julia E},
  booktitle={Joint European Conference on Machine Learning and Knowledge Discovery in Databases},
  pages={447--462},
  year={2025},
  organization={Springer}
}

@article{hu2024gcc,
  title={GCC: Generative Calibration Clustering},
  author={Xia, Haifeng and Huang, Hai and Ding, Zhengming},
  journal={arXiv preprint arXiv:2404.09115},
  year={2024}
}

@article{lecun1998gradient,
  title={Gradient-based learning applied to document recognition},
  author={LeCun, Yann and Bottou, L{\'e}on and Bengio, Yoshua and Haffner, Patrick},
  journal={Proceedings of the IEEE},
  volume={86},
  number={11},
  pages={2278--2332},
  year={1998},
  publisher={Ieee}
}

@article{xiao2017fashion,
  title={Fashion-MNIST: a Novel Image Dataset for Benchmarking Machine Learning Algorithms},
  author={Xiao, Han and Rasul, Kashif and Vollgraf, Roland},
  journal={arXiv preprint arXiv:1708.07747},
  year={2017}
}

@article{lamontagne2019oasis,
  title={OASIS-3: longitudinal neuroimaging, clinical, and cognitive dataset for normal aging and Alzheimer disease},
  author={LaMontagne, Pamela J and Benzinger, Tammie LS and Morris, John C and Keefe, Sarah and Hornbeck, Russ and Xiong, Chengjie and Grant, Elizabeth and Hassenstab, Jason and Moulder, Krista and Vlassenko, Andrei G and others},
  journal={medrxiv},
  pages={2019--12},
  year={2019},
  publisher={Cold Spring Harbor Laboratory Press}
}

@article{dalca2019learning,
  title={Learning conditional deformable templates with convolutional networks},
  author={Dalca, Adrian and Rakic, Marianne and Guttag, John and Sabuncu, Mert},
  journal={Advances in neural information processing systems},
  volume={32},
  year={2019}
}

@article{ho2020denoising,
  title={Denoising diffusion probabilistic models},
  author={Ho, Jonathan and Jain, Ajay and Abbeel, Pieter},
  journal={Advances in neural information processing systems},
  volume={33},
  pages={6840--6851},
  year={2020}
}

@article{ba2013convergence,
  title={Convergence and stability of iteratively re-weighted least squares algorithms},
  author={Ba, Demba and Babadi, Behtash and Purdon, Patrick L and Brown, Emery N},
  journal={IEEE Transactions on Signal Processing},
  volume={62},
  number={1},
  pages={183--195},
  year={2013},
  publisher={IEEE}
}

@article{kingma2013auto,
  title={Auto-encoding variational bayes},
  author={Kingma, Diederik P and Welling, Max},
  journal={arXiv preprint arXiv:1312.6114},
  year={2013}
}

@article{mclachlan2019finite,
  title={Finite mixture models},
  author={McLachlan, Geoffrey J and Lee, Sharon X and Rathnayake, Suren I},
  journal={Annual review of statistics and its application},
  volume={6},
  number={1},
  pages={355--378},
  year={2019},
  publisher={Annual Reviews}
}

@inproceedings{abiri2020variational,
  title={Variational auto-encoders with Student’s t-prior},
  author={Abiri, Najmeh and Ohlsson, Mattias},
  booktitle={27th European Symposium on Artificial Neural Networks, Computational Intelligence and Machine Learning},
  year={2019}
}

@inproceedings{van2020scan,
  title={Scan: Learning to classify images without labels},
  author={Van Gansbeke, Wouter and Vandenhende, Simon and Georgoulis, Stamatios and Proesmans, Marc and Van Gool, Luc},
  booktitle={European conference on computer vision},
  pages={268--285},
  year={2020},
  organization={Springer}
}

@inproceedings{wang2023infodiffusion,
  title={Infodiffusion: Representation learning using information maximizing diffusion models},
  author={Wang, Yingheng and Schiff, Yair and Gokaslan, Aaron and Pan, Weishen and Wang, Fei and De Sa, Christopher and Kuleshov, Volodymyr},
  booktitle={International conference on machine learning},
  pages={36336--36354},
  year={2023},
  organization={PMLR}
}

@article{svensen2005robust,
  title={Robust Bayesian mixture modelling},
  author={Svens{\'e}n, Markus and Bishop, Christopher M},
  journal={Neurocomputing},
  volume={64},
  pages={235--252},
  year={2005},
  publisher={Elsevier}
}

@inproceedings{boenninghoff2020variational,
  title     = {Variational Autoencoder with Embedded {Student-$t$} Mixture Model for Authorship Attribution},
  author    = {Boenninghoff, Benedikt and Zeiler, Steffen and Nickel, Robert and Kolossa, Dorothea},
  booktitle = {Proceedings of the 28th International Conference on Computational Linguistics (COLING)},
  pages     = {519--529},
  year      = {2020},
  address   = {Barcelona, Spain (Online)},
  publisher = {International Committee on Computational Linguistics},
  doi       = {10.18653/v1/2020.coling-main.45}
}

@inproceedings{wang2021bayesian,
  title     = {{Bayesian} Atlas Building with Hierarchical Priors for Subject-specific Regularization},
  author    = {Wang, Jian and Zhang, Miaomiao},
  booktitle = {Medical Image Computing and Computer Assisted Intervention (MICCAI)},
  year      = {2021},
  publisher = {Springer},
  doi       = {10.1007/978-3-030-87202-1_8}
}

@article{wang2022geo,
  title={Geo-sic: learning deformable geometric shapes in deep image classifiers},
  author={Wang, Jian and Zhang, Miaomiao},
  journal={Advances in Neural Information Processing Systems},
  volume={35},
  pages={27994--28007},
  year={2022}
}

@inproceedings{xing2026manifold,
  title={On-Manifold Variational Learning with Heat-Kernel Priors},
  author={Xing, Jiarui and Zeevi, Tal and Wu, Nian and Wang, Jian},
  booktitle={International Conference on Medical Image Computing and Computer-Assisted Intervention},
  pages={110--120},
  year={2026},
  organization={Springer}
}

\clearpage
\appendix

\numberwithin{equation}{section}
\numberwithin{figure}{section}
\numberwithin{table}{section}
\numberwithin{proposition}{section}
\numberwithin{remark}{section}

\section*{Appendix}
This appendix supplies the complete analytical derivations underlying
the closed-form ELBO presented in the main paper (Sections~\ref{app:recap}--\ref{app:full-cov}), and additional experimental results (Section~\ref{app:vis}). The central claim
of the paper, that the Student-$t$ variant of VaDE admits an exact
analytical objective with no approximation in any KL term, rests on
three results: the scale-mixture identity that recasts the
Student-$t$ prior as a Gamma--Gaussian hierarchy
(Section~\ref{app:proof-smg}), the Gamma--Gaussian conjugacy that
determines the optimal scale posterior in closed form
(Section~\ref{app:aux-posterior}), and the analytical integration of
the auxiliary variable that collapses the augmented ELBO into three
tractable KL divergences (Section~\ref{app:elbo-derivation}). We
prove each in turn and extend the result to the full-covariance
setting (Section~\ref{app:full-cov}). Section~\ref{app:recap}
collects the notation and the key equations from the main paper that
serve as reference points for the derivations. 

\section{Notation and Key Equations}\label{app:recap}

For self-containedness, we restate the conventions and the principal
equations from the main paper. 

\paragraph{Notation.} Observations $\bx\in\R^D$ are embedded into
latent codes $\bz\in\R^d$ with cluster labels $c\in\{1,\dots,K\}$. We
write $\N(\bmu,\bm{\Lambda}^{-1})$ for a Gaussian with mean $\bmu$
and precision $\bm{\Lambda}$, $\St_\nu(\bmu,\bm{\Lambda}^{-1})$ for a
Student's $t$ with $\nu$ degrees of freedom, and $\Gam(\alpha,\beta)$
for a Gamma in shape--rate form (mean $\alpha/\beta$). Cluster
precisions $\bm{\Lambda}_c$ are diagonal in the main text with
entries $\{\Lambda_{c,j}\}_{j=1}^d$; the full-covariance case is
treated in Section~\ref{app:full-cov}. The digamma function is
$\psi$.

\paragraph{Generative model.}
\begin{equation}\label{app:gen}
c\sim\mathrm{Cat}(\bm{\pi}),\quad
u\mid c\sim\Gam(\nu_c/2,\nu_c/2),\quad
\bz\mid u,c\sim\N(\bmu_c,(u\bm{\Lambda}_c)^{-1}),\quad
\bx\mid\bz\sim p_\theta(\bx\mid\bz).
\end{equation}

\paragraph{Variational family.}
\begin{equation}\label{app:q-fact}
q(\bz,u,c\mid\bx)
=q_\phi(\bz\mid\bx)\,q(u\mid\bx,c)\,q(c\mid\bx),
\end{equation}
with
$q_\phi(\bz\mid\bx)=\N(\bmu_\phi(\bx),\mathrm{diag}(\bm{\sigma}_\phi^2(\bx)))$
and $q(c\mid\bx)=\gamma_c$.

\paragraph{Encoder-averaged Mahalanobis distance.}
\begin{equation}\label{app:delta-bar}
\bar\Delta_c^2
\coloneqq\E_{q_\phi(\bz\mid\bx)}\!\bigl[
(\bz-\bmu_c)^\top\bm{\Lambda}_c(\bz-\bmu_c)\bigr]
=\sum_{j=1}^d\Lambda_{c,j}\bigl[\sigma_{\phi,j}^2
+(\mu_{\phi,j}-\mu_{c,j})^2\bigr].
\end{equation}

\paragraph{Optimal scale posterior.}
\begin{equation}\label{app:u-posterior}
q^\star(u\mid\bx,c)=\Gam(\tilde\alpha_c,\tilde\beta_c),\quad
\tilde\alpha_c=\tfrac{\nu_c+d}{2},\quad
\tilde\beta_c=\tfrac{\nu_c+\bar\Delta_c^2}{2}.
\end{equation}

\paragraph{Sufficient statistics.}
\begin{equation}\label{app:u-moments}
\E[u]=\tfrac{\nu_c+d}{\nu_c+\bar\Delta_c^2},\qquad
\E[\log u]=\psi(\tilde\alpha_c)-\log\tilde\beta_c.
\end{equation}

\paragraph{Collapsed ELBO and KL terms.}
\begin{align}
\mathcal{L}(\bx)&=\E_{q_\phi}[\log p_\theta(\bx\mid\bz)]
-\sum_{c=1}^K\gamma_c\bigl[
\widetilde{\KL}^{(c)}_{\text{Gauss}}
+\KL^{(c)}_{\text{Gamma}}
+\KL^{(c)}_{\text{Cat}}\bigr],\label{app:our-elbo}\\[2pt]
\widetilde{\KL}^{(c)}_{\text{Gauss}}&=
-\tfrac{1}{2}\!\left[d+d\,\E[\log u]+\log|\bm{\Lambda}_c|
+\textstyle\sum_j\log\sigma_{\phi,j}^2-\E[u]\,\bar\Delta_c^2\right],
\label{app:kl-gauss}\\[2pt]
\KL^{(c)}_{\text{Gamma}}&=
(\tilde\alpha_c-\alpha_0)\,\psi(\tilde\alpha_c)
+\log\Gamma(\alpha_0)-\log\Gamma(\tilde\alpha_c)
+\alpha_0\log\tfrac{\tilde\beta_c}{\beta_0}
+\tilde\alpha_c\,\tfrac{\beta_0-\tilde\beta_c}{\tilde\beta_c},
\label{app:kl-gamma}
\end{align}
with prior $\alpha_0=\beta_0=\nu_c/2$, and
$\KL^{(c)}_{\text{Cat}}=\log\gamma_c-\log\pi_c$.

\section{Proof of the Scale Mixture Representation}\label{app:proof-smg}

The starting point of our derivation is the observation that a
Student-$t$ density is itself an integral of Gaussians against a
Gamma. This identity is what allows us to introduce the auxiliary
variable $u$ in~\eqref{app:gen} without changing the marginal
distribution of $\bz$, and is therefore the structural foundation on
which the subsequent collapsed inference rests.

\begin{proposition}[Scale Mixture of Gaussians]
\label{prop:smg-sm}
If $u\sim\Gam(\nu/2,\nu/2)$ and
$\bz\mid u\sim\N(\bmu,(u\bm{\Lambda})^{-1})$, then
$\bz\sim\St_\nu(\bmu,\bm{\Lambda}^{-1})$.
\end{proposition}

\begin{proof}
The proof proceeds by direct integration of the joint density
$p(\bz,u)=p(\bz\mid u)p(u)$ over $u$. The key step is recognising
that the integrand is, up to constants, a Gamma kernel.

\paragraph{Step 1: joint density.}
The conditional Gaussian is
\(p(\bz\mid u)=\frac{|u\bm{\Lambda}|^{1/2}}{(2\pi)^{d/2}}
\exp\{-\frac{u}{2}\Delta^2\}\)
with $\Delta^2=(\bz-\bmu)^\top\bm{\Lambda}(\bz-\bmu)$ and
$|u\bm{\Lambda}|=u^d|\bm{\Lambda}|$. The Gamma prior is
\(p(u)=\frac{(\nu/2)^{\nu/2}}{\Gamma(\nu/2)}u^{\nu/2-1}e^{-\nu u/2}\).
Multiplying these and collecting powers of $u$ produces an integrand
proportional to $u^{(\nu+d)/2-1}\exp\{-\frac{u}{2}(\nu+\Delta^2)\}$,
which is the kernel of a Gamma distribution.

\paragraph{Step 2: integration.}
\begin{equation}\label{app:integral}
p(\bz)=\frac{|\bm{\Lambda}|^{1/2}(\nu/2)^{\nu/2}}{(2\pi)^{d/2}\Gamma(\nu/2)}
\int_0^\infty u^{(\nu+d)/2-1}\exp\!\left\{-\tfrac{u}{2}(\nu+\Delta^2)\right\}\dif u.
\end{equation}
The remaining integral is in standard form
$\int_0^\infty t^{a-1}e^{-bt}\dif t=\Gamma(a)/b^a$, with
$a=(\nu+d)/2$ and $b=(\nu+\Delta^2)/2$, giving
\begin{equation}
\int_0^\infty u^{(\nu+d)/2-1}\exp\!\left\{-\tfrac{u}{2}(\nu+\Delta^2)\right\}\dif u
=\frac{\Gamma\!\bigl(\frac{\nu+d}{2}\bigr)}
       {\bigl(\frac{\nu+\Delta^2}{2}\bigr)^{(\nu+d)/2}}.
\end{equation}

\paragraph{Step 3: simplification.}
Using
$\bigl((\nu+\Delta^2)/2\bigr)^{(\nu+d)/2}
=(\nu/2)^{(\nu+d)/2}(1+\Delta^2/\nu)^{(\nu+d)/2}$,
$(\nu/2)^{\nu/2}/(\nu/2)^{(\nu+d)/2}=(\nu/2)^{-d/2}$, and
$(\nu/2)^{-d/2}/(2\pi)^{d/2}=(\nu\pi)^{-d/2}$, the constants combine
into the standard Student-$t$ normalising factor:
\begin{equation}\label{app:final-t}
p(\bz)=\frac{\Gamma\!\bigl(\frac{\nu+d}{2}\bigr)|\bm{\Lambda}|^{1/2}}
            {\Gamma\!\bigl(\frac{\nu}{2}\bigr)(\nu\pi)^{d/2}}
\left[1+\tfrac{\Delta^2}{\nu}\right]^{-(\nu+d)/2},
\end{equation}
which is the density of $\St_\nu(\bmu,\bm{\Lambda}^{-1})$.
\end{proof}

The identity has two consequences that the main paper relies on.
First, the augmented model~\eqref{app:gen} is statistically
indistinguishable from a $t$-mixture VaDE: every marginal
distribution is unchanged. Second, conditional on $u$, the prior on
$\bz$ is Gaussian, which is precisely what makes the inference
problem tractable in the next section.

\section{Optimal Scale Posterior}\label{app:aux-posterior}

The structured factorisation~\eqref{app:q-fact} leaves the form of
$q(u\mid\bx,c)$ unspecified. We now show that this freedom is in
fact illusory: the coordinate-ascent optimum is uniquely determined
by Gamma--Gaussian conjugacy, and admits the closed-form expression
quoted in~\eqref{app:u-posterior}. This result is what allows us to
treat $u$ as analytically integrable rather than approximated.

\begin{proposition}[Optimal scale posterior]
\label{prop:opt-q}
Under the augmented model~\eqref{app:gen} and the variational
family~\eqref{app:q-fact}, the optimal $q(u\mid\bx,c)$ is the Gamma
distribution given by~\eqref{app:u-posterior}.
\end{proposition}

\begin{proof}
The coordinate-ascent variational inference (CAVI) update for
$q(u\mid\bz,c)$ holding $q_\phi(\bz\mid\bx)$ and $q(c\mid\bx)$ fixed
is given by Bayes' rule restricted to the conditional model:
\begin{equation}
q(u\mid\bz,c)\propto p(\bz\mid u,c)\,p(u\mid c).
\end{equation}
Substituting the conditional Gaussian and Gamma densities
from~\eqref{app:gen},
\begin{equation}
q(u\mid\bz,c)\propto u^{d/2}e^{-\frac{u}{2}\Delta_c^2(\bz)}
\cdot u^{\nu_c/2-1}e^{-\frac{\nu_c}{2}u}
=u^{(\nu_c+d)/2-1}e^{-\frac{u}{2}(\nu_c+\Delta_c^2(\bz))},
\end{equation}
which is recognised as the kernel of
$\Gam\bigl((\nu_c+d)/2,\,(\nu_c+\Delta_c^2(\bz))/2\bigr)$. Taking
the expectation under the encoder $q_\phi(\bz\mid\bx)$ replaces
$\Delta_c^2(\bz)$ by its mean $\bar\Delta_c^2$
from~\eqref{app:delta-bar}, yielding
$q^\star(u\mid\bx,c)=\Gam(\tilde\alpha_c,\tilde\beta_c)$
with the parameters in~\eqref{app:u-posterior}.
\end{proof}

The form of $\tilde\alpha_c$ and $\tilde\beta_c$ already foreshadows
the robustness mechanism analysed in the main paper: an inlier with
$\bar\Delta_c^2\approx d$ produces a Gamma centred near the prior
mean and hence $\E[u]\approx 1$, whereas an outlier with
$\bar\Delta_c^2\gg\nu_c$ produces a Gamma whose mass concentrates at
small $u$, and hence $\E[u]\to 0$. Both expectations are required
for the collapsed ELBO derived next, and both are available in
closed form via~\eqref{app:u-moments}.

\section{Collapsed ELBO Derivation}\label{app:elbo-derivation}

We now derive the collapsed ELBO~\eqref{app:our-elbo}. The
derivation proceeds in three stages: (i) write the augmented ELBO
under the variational family~\eqref{app:q-fact}; (ii) regroup terms
into three KL divergences corresponding to $\bz$, $u$, and $c$; (iii)
substitute the optimal $q^\star(u\mid\bx,c)$ and integrate $u$
analytically. Each KL term then reduces to a closed-form expression
in the sufficient statistics~\eqref{app:u-moments}.

\paragraph{Stage 1: augmented ELBO.}
By the definition of the ELBO under~\eqref{app:gen}
and~\eqref{app:q-fact},
\begin{align}
\mathcal{L}=\E_q\!\bigl[
&\log p_\theta(\bx\mid\bz)
+\log p(\bz\mid u,c)+\log p(u\mid c)+\log p(c)\notag\\
&-\log q_\phi(\bz\mid\bx)-\log q(u\mid\bz,c)-\log q(c\mid\bx)
\bigr].\label{app:elbo-expand}
\end{align}

\paragraph{Stage 2: regrouping into three KL divergences.}
Pairing each prior term with its variational counterpart and using
$q(c\mid\bx)=\gamma_c$,
\begin{align}
\mathcal{L}&=\E_{q_\phi}[\log p_\theta(\bx\mid\bz)]
-\sum_c\gamma_c\!\left[
\underbrace{\E_{q_\phi,q_u}\!\bigl[\log\tfrac{q_\phi(\bz\mid\bx)}{p(\bz\mid u,c)}\bigr]}_{\widetilde{\KL}_{\text{Gauss}}^{(c)}}
+\underbrace{\E_{q_\phi}\!\bigl[\KL[q(u\mid\bz,c)\|p(u\mid c)]\bigr]}_{\KL_{\text{Gamma}}^{(c)}}
+\underbrace{\log\tfrac{\gamma_c}{\pi_c}}_{\KL_{\text{Cat}}^{(c)}}
\right].\label{app:elbo-3kl}
\end{align}
Each of the three braced terms is now treated separately.

\paragraph{Stage 3a: collapsing the Gaussian KL.}
Substituting the Gaussian log-density
$\log p(\bz\mid u,c)=-\tfrac{d}{2}\log 2\pi+\tfrac{d}{2}\log u
+\tfrac{1}{2}\log|\bm{\Lambda}_c|-\tfrac{u}{2}\Delta_c^2(\bz)$ and the
encoder log-density into the first term, and using the four
identities
\begin{equation}
\E_{q_u}[\log u]=\E[\log u],\quad
\E_{q_u}[u]=\E[u],\quad
\E_{q_\phi}[\Delta_c^2(\bz)]=\bar\Delta_c^2,\quad
\E_{q_\phi}\!\Bigl[\sum_j\tfrac{(z_j-\mu_{\phi,j})^2}{\sigma_{\phi,j}^2}\Bigr]=d,
\end{equation}
the entire $u$-dependence collapses into the two scalars $\E[u]$ and
$\E[\log u]$:
\begin{align}
\widetilde{\KL}_{\text{Gauss}}^{(c)}
&=\bigl(-\tfrac{1}{2}\textstyle\sum_j\log\sigma_{\phi,j}^2-\tfrac{d}{2}\bigr)
-\bigl(\tfrac{d}{2}\E[\log u]+\tfrac{1}{2}\log|\bm{\Lambda}_c|
-\tfrac{\E[u]}{2}\bar\Delta_c^2\bigr)\notag\\
&=-\tfrac{1}{2}\bigl[d+d\,\E[\log u]+\log|\bm{\Lambda}_c|
+\textstyle\sum_j\log\sigma_{\phi,j}^2-\E[u]\,\bar\Delta_c^2\bigr],
\end{align}
which is~\eqref{app:kl-gauss}.

\paragraph{Stage 3b: closing the Gamma KL.}
The second term is the KL divergence between two Gamma distributions
in shape--rate form, $q=\Gam(\tilde\alpha,\tilde\beta)$ and
$p=\Gam(\alpha_0,\beta_0)$. Using
$\E_q[\log u]=\psi(\tilde\alpha)-\log\tilde\beta$ and
$\E_q[u]=\tilde\alpha/\tilde\beta$ in the standard Gamma--Gamma KL
formula,
\begin{equation}
\KL[q\|p]=(\tilde\alpha-\alpha_0)\,\psi(\tilde\alpha)
+\log\Gamma(\alpha_0)-\log\Gamma(\tilde\alpha)
+\alpha_0\log\tfrac{\tilde\beta}{\beta_0}
+\tilde\alpha\,\tfrac{\beta_0-\tilde\beta}{\tilde\beta},
\end{equation}
which becomes~\eqref{app:kl-gamma} on substituting
$\tilde\alpha=\tilde\alpha_c$, $\tilde\beta=\tilde\beta_c$, and
$\alpha_0=\beta_0=\nu_c/2$.

\paragraph{Stage 3c: categorical KL.}
The third term is identical to that of Gaussian VaDE,
$\KL_{\text{Cat}}^{(c)}=\log\gamma_c-\log\pi_c$, since the
augmentation does not alter the cluster prior.

\paragraph{Recovery of the Gaussian limit.}
As $\nu_c\to\infty$, the Gamma prior on $u$ concentrates at $1$,
giving $\E[u]\to 1$ and $\E[\log u]\to 0$, while
$\KL_{\text{Gamma}}^{(c)}\to 0$. Substituting into the three KL
expressions, $\widetilde{\KL}_{\text{Gauss}}^{(c)}$ reduces to the
standard Gaussian--Gaussian KL of VaDE and the Gamma term vanishes
identically. The collapsed ELBO~\eqref{app:our-elbo} therefore
recovers the Gaussian VaDE objective as a strict limiting case.

\paragraph{Summary.}
Each KL term in~\eqref{app:our-elbo} has been derived in closed form,
without recourse to Wilson--Hilferty transformations, moment matching,
or variational mixture bounds. The auxiliary variable $u$ enters the
final loss only through its first two moments $\E[u]$ and $\E[\log u]$,
both available in closed form via~\eqref{app:u-moments}, so no
sampling of $u$ is required at any point during training.

\section{Full-Covariance Extension}\label{app:full-cov}

The main paper assumes diagonal cluster precisions $\bm{\Lambda}_c$
to match the diagonal Gaussian encoder. We show here that the
derivation extends to the full-covariance setting without
modification of any KL term, so that the closed-form result is not
an artefact of the diagonal assumption.

For full $\bm{\Lambda}_c\in\R^{d\times d}$, the encoder-averaged
Mahalanobis distance generalises to
\begin{equation}\label{app:full-cov-delta}
\bar\Delta_c^2
=\tr(\bm{\Lambda}_c\bm{\Sigma}_\phi)
+(\bmu_\phi-\bmu_c)^\top\bm{\Lambda}_c(\bmu_\phi-\bmu_c),
\end{equation}
where $\bm{\Sigma}_\phi=\mathrm{diag}(\bm{\sigma}_\phi^2)$ for the
diagonal encoder used in the main text (the result also holds for a
full-covariance encoder, replacing $\bm{\Sigma}_\phi$ with the
encoder covariance), and $\log|\bm{\Lambda}_c|$ is the log-determinant
of the full precision matrix.

The derivations of the optimal scale posterior~\eqref{app:u-posterior},
the sufficient statistics~\eqref{app:u-moments}, and the
KL terms~\eqref{app:kl-gauss}--\eqref{app:kl-gamma} are unchanged in
form: each depends on $\bm{\Lambda}_c$ only through the scalar
$\bar\Delta_c^2$ and the determinant $\log|\bm{\Lambda}_c|$. The
Gamma--Gaussian conjugacy underlying Proposition~\ref{prop:opt-q}
holds for any positive-definite $\bm{\Lambda}_c$, and the
Stage-3a collapse goes through with the substitution
$\sum_j\log\sigma_{\phi,j}^2\to\log|\bm{\Sigma}_\phi|$ for a
full-covariance encoder. The closed-form character of the ELBO is
therefore preserved.

\rev{
\section{Related Experimental Results}\label{app:vis}
\subsection{Visualization}
We present visualization results in Figure~\ref{fig:noisy_samples} for different types of noise injection on the MNIST, Fashion MNIST, and OASIS datasets. We provide the detailed experiments of three 2-d views of OASIS in Figure~\ref{fig:oasis_3view}.

\begin{figure}[!htbp]
  \centering
  \includegraphics[width=\linewidth]{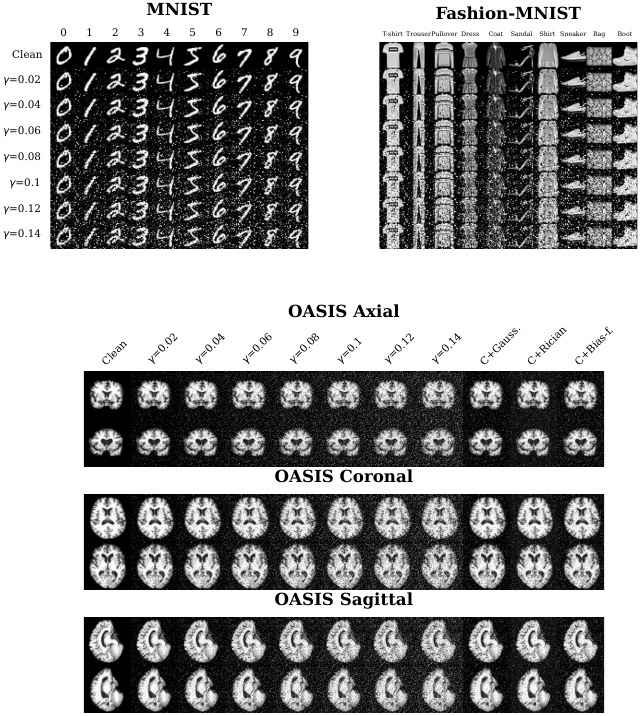}
  \caption{Noisy input visualization across all benchmark datasets.
  \textbf{Top two panels:} MNIST and Fashion-MNIST with one sample per class (columns) under clean input and seven Cauchy noise levels $\gamma \in \{0.02, 0.04, \ldots, 0.14\}$ (rows).
  \textbf{Bottom three panels:} OASIS brain MRI in coronal, axial, and sagittal views (2 representative slices per panel).
  Each column corresponds to a noise condition: clean, seven Cauchy levels, and three composite types: Cauchy+Gaussian, Cauchy+Rician, and Cauchy+Bias-field at medium strength.
  Cauchy noise produces heavy-tailed outlier pixels that progressively corrupt local structure while global anatomy remains recognizable.}
  \label{fig:noisy_samples}
\end{figure}

\begin{figure}[!htbp]
  \centering
  \begin{subfigure}[b]{\linewidth}
    \centering
    \includegraphics[width=\linewidth]{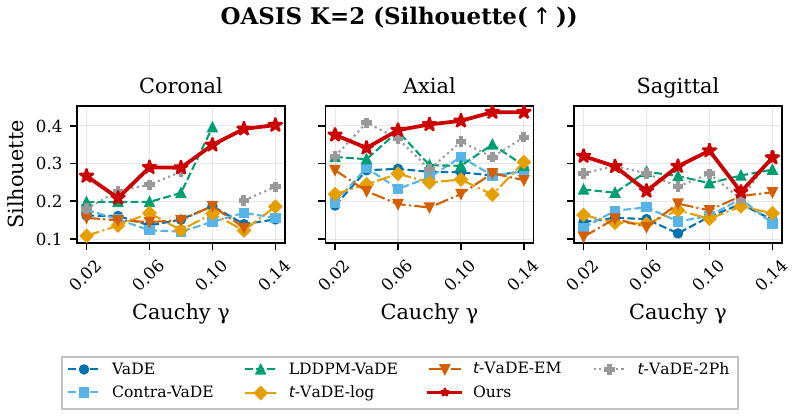}
    \caption{Silhouette score across three views.}
  \end{subfigure}
  \begin{subfigure}[b]{\linewidth}
    \centering
    \includegraphics[width=\linewidth]{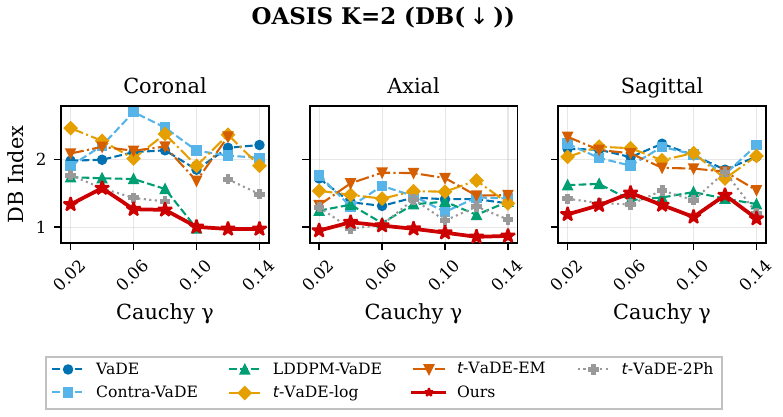}
    \caption{Davies--Bouldin index across three views.}
  \end{subfigure}
  \caption{OASIS ($K{=}2$) clustering quality under Cauchy noise for coronal, axial, and sagittal views, respectively.
  Our method consistently achieves the highest Silhouette score and lowest Davies--Bouldin index across all three anatomical views and noise levels (consistent with the results in Table~\ref{tab:main} in the main paper), demonstrating robust clustering quality regardless of the imaging plane.}
  \label{fig:oasis_3view}
\end{figure}

\subsection{Computational Costs}\label{app:runtime}
We distinguish training-update cost from end-to-end cluster assignment.
Training measurements use an NVIDIA RTX 5880 Ada GPU, batch size 32,
and 101 timed optimizer updates per reported configuration.
The two-phase baseline is timed in its second training phase.
These measurements do not represent the total duration of 101 epochs.

\begin{table}[htbp]
\centering
\caption{Controlled training cost in ms/update.}
\label{tab:runtime-training}
\begin{tabular}{lrr}
\toprule
Method & MNIST & Fashion-MNIST \\
\midrule
VaDE & 27.74 & 27.05 \\
Contra-VaDE & 28.79 & 29.14 \\
LDDPM-VaDE & 31.99 & 30.20 \\
$t$-VaDE-log & 28.48 & 30.89 \\
$t$-VaDE-EM & 29.14 & 30.00 \\
Ours & 33.21 & 32.00 \\
$t$-VaDE-2Ph & 24.53 & 25.66 \\
\bottomrule
\end{tabular}
\end{table}

For the reverse-inference benchmark, existing checkpoints are evaluated
at Cauchy noise $\gamma=0.14$ over three seeds. The encoded latent is forward-noised to the final diffusion time, followed by 100 deterministic DDIM reverse updates before mixture assignment. The original raw-latent assignment path is not used in this benchmark.
VaDE and our method use their usual encode-and-assign paths. Latency includes encoding, reverse updates where applicable, and mixture assignment. We use batch size 256, discard five warm-up calls, and average 20 CUDA-event timing repetitions per seed. Training still samples one diffusion time per update; the network is not retrained for this timing experiment.

\begin{table}[htbp]
\centering
\caption{End-to-end assignment latency in ms/sample.
The $33\times$/$36\times$ comparison is against the specified
100-step reverse-inference path.}
\label{tab:runtime-assignment}
\begin{tabular}{lrrrr}
\toprule
Method & Reverse calls & MNIST & Fashion & Params (MB) \\
\midrule
VaDE & 0 & 0.0036 & 0.0036 & 6.95 \\
LDDPM-VaDE-100 & 100 & 0.1422 & 0.1422 & 6.98 \\
Ours & 0 & 0.0043 & 0.0040 & 6.95 \\
\bottomrule
\end{tabular}
\end{table}
}

\rev{\subsection{Diagonal versus Full Precision}\label{app:cov-ablation}
We compare diagonal and full cluster precision matrices at Cauchy noise $\gamma=0.14$ over three paired seeds, while retaining the same encoder and other settings. Full precision preserves the analytic bound and increases computation from $O(BKd)$ to $O(BKd^2)$ for the mixture terms.

\begin{table}[htbp]
\centering
\caption{Diagonal/full precision ablation at Cauchy $\gamma=0.14$. ACC is mean$\pm$sample SD across three matched seeds. Time is measured in ms/update.}
\label{tab:cov-ablation}
\begin{tabular}{lrrrr}
\toprule
Dataset & Diag. ACC & Full ACC & Diag. time & Full time \\
\midrule
MNIST & $.796\pm.044$ & $.812\pm.075$ & 33.21 & 34.22 \\
Fashion-MNIST & $.543\pm.007$ & $.568\pm.016$ & 32.00 & 33.66 \\
\bottomrule
\end{tabular}
\end{table}

The full model therefore provides an analytic option when cross-dimensional dependencies are useful, while diagonal precision remains the lower-cost default.}

\rev{\subsection{Multi-Seed Results and Significance}\label{app:multiseed}
We performance experiments over five random seeds at three Cauchy-noise scales. The mean paired difference on MNIST is 0.079 (95\% CI [0.052, 0.106]) against $t$-VaDE-log; on Fashion-MNIST it is 0.116 (95\% CI [0.032, 0.200]) against $t$-VaDE-EM. Paired $t$-tests with Holm correction across these two dataset-level comparisons yield $p=0.0025$ and $p=0.018$.
\begin{table}[!htb]
\centering
\caption{MNIST ACC over five seeds. Clean inputs are excluded.}
\label{tab:mnist-five-seed}
\begin{tabular}{lccc}
\toprule
Method & $\gamma=.02$ & $\gamma=.08$ & $\gamma=.14$ \\
\midrule
VaDE & $.663\pm.051$ & $.610\pm.055$ & $.622\pm.042$ \\
Contra-VaDE & $.694\pm.041$ & $.645\pm.049$ & $.606\pm.035$ \\
LDDPM-VaDE & $.328\pm.096$ & $.272\pm.102$ & $.197\pm.070$ \\
$t$-VaDE-log & $.749\pm.022$ & $.713\pm.030$ & $.687\pm.037$ \\
$t$-VaDE-EM & $.636\pm.030$ & $.627\pm.051$ & $.659\pm.040$ \\
$t$-VaDE-2Ph & $.652\pm.061$ & $.697\pm.045$ & $.625\pm.068$ \\
Ours & $.817\pm.045$ & $.778\pm.010$ & $.792\pm.037$ \\
\bottomrule
\end{tabular}
\end{table}

\begin{table}[!htb]
\centering
\caption{Fashion-MNIST ACC over five seeds. Clean inputs are excluded.}
\label{tab:fashion-five-seed}
\begin{tabular}{lccc}
\toprule
Method & $\gamma=.02$ & $\gamma=.08$ & $\gamma=.14$ \\
\midrule
VaDE & $.455\pm.136$ & $.424\pm.123$ & $.311\pm.131$ \\
Contra-VaDE & $.479\pm.028$ & $.380\pm.153$ & $.356\pm.070$ \\
LDDPM-VaDE & $.191\pm.063$ & $.237\pm.077$ & $.185\pm.027$ \\
$t$-VaDE-log & $.502\pm.025$ & $.407\pm.119$ & $.363\pm.117$ \\
$t$-VaDE-EM & $.505\pm.026$ & $.483\pm.074$ & $.317\pm.168$ \\
$t$-VaDE-2Ph & $.516\pm.023$ & $.415\pm.144$ & $.261\pm.205$ \\
Ours & $.552\pm.026$ & $.555\pm.019$ & $.548\pm.010$ \\
\bottomrule
\end{tabular}
\end{table}}

\rev{\subsection{Degrees-of-Freedom Initialization}\label{app:nu-init}
We test initialization of $\nu_0\in\{2,4,5,8,16,32\}$ and evaluate three runs per setting at Cauchy $\gamma=0.14$. Entries are mean$\pm$sample SD.

\begin{table}[htbp]
\centering
\caption{Sensitivity to the initial degrees of freedom.}
\label{tab:nu-init}
\setlength{\tabcolsep}{5.5pt}
\begin{tabular}{lcccccc}
\toprule
$\nu_0$ & 2 & 4 & 5 & 8 & 16 & 32 \\
\midrule
MNIST
& $.830\pm.044$ & $.835\pm.073$ & $.796\pm.044$
& $.852\pm.049$ & $.803\pm.015$ & $.768\pm.039$ \\
Fashion
& $.548\pm.023$ & $.544\pm.009$ & $.543\pm.007$
& $.536\pm.027$ & $.562\pm.017$ & $.551\pm.050$ \\
\bottomrule
\end{tabular}
\end{table}

The Fashion-MNIST means range from 0.536 to 0.562. MNIST varies more, from 0.768 to 0.852, with the lowest mean at $\nu_0=32$.}

\clearpage

\section*{NeurIPS Paper Checklist}

\begin{enumerate}

\item {\bf Claims}
    \item[] Question: Do the main claims made in the abstract and introduction accurately reflect the paper's contributions and scope?
    \item[] Answer: \answerYes{}
    \item[] Justification: The abstract and introduction state four claims: (i) a fully closed-form ELBO for a Student's $t$-mixture deep clustering model, (ii) inherent noise immunity through implicit IRLS reweighting, (iii) state-of-the-art clustering accuracy and stability on standard benchmarks, and (iv) clinical utility on a real neurodegenerative disease cohort. 
    
\item {\bf Limitations}
    \item[] Answer: \answerYes{}
    \item[] Justification: Section~\ref{sec:conclusion} discusses the
elliptical component geometry and the fixed number of components $K$.
Appendix~\ref{app:nu-init} reports sensitivity to the initial
degrees of freedom.

\item {\bf Theory assumptions and proofs}
    \item[] Answer: \answerYes{}
    \item[] Justification: All theoretical claims are stated with explicit assumptions in Section~\ref{sec:method}. The closed-form ELBO~\eqref{eq:our-elbo} is derived from the structured variational posterior \eqref{eq:variational} together with the optimal $q^\star(u\mid\bx,c)$ given by~\eqref{eq:q-u-star}, with all expectation identities and the extension to full-covariance $\bm{\Lambda}_c$ provided in Appendix.

\item {\bf Experimental result reproducibility}
    \item[] Answer: \answerYes{}
    \item[] Justification: Section~\ref{subsec:exp-setup} describes the encoder--decoder architecture, optimizer, learning rate, batch size, number of training epochs, initialization scheme for $(\bmu_c, \bm{\Lambda}_c, \pi_c, \nu_c)$, and the noise corruption protocol. Appendix~\ref{app:vis} provides supplementary experimental results and timing protocols; training and evaluation scripts are available in the public code repository.

\item {\bf Open access to data and code}
    \item[] Answer: \answerYes{}
    \item[] Justification: Our source code is publicly available at
\url{https://github.com/xuzhikangnba/torch-tvae}. The repository includes the Student's $t$ VAE implementation, the original multi-component models, training and evaluation scripts, installation instructions, and reproduction reports.
MNIST and Fashion-MNIST can be downloaded through the provided training pipeline. OASIS data must be obtained separately from the dataset provider under its access conditions.

\item {\bf Experimental setting/details}
    \item[] Answer: \answerYes{}
    \item[] Justification: Section~\ref{subsec:exp-setup} specifies the shared backbone, AdamW optimizer, learning rate, weight decay, batch size, training duration, and softplus parameterization of $\nu_c$. Appendix~\ref{app:nu-init} reports the initialization sensitivity study.

\item {\bf Experiment statistical significance}
    \item[] Answer: \answerYes{}
    \item[] Justification: Section~\ref{subsec:exp-clustering} reports an additional five-seed comparison on MNIST and Fashion-MNIST, including 95\% confidence intervals and $p$-values. This analysis is separate from the main results aggregated over noise levels or anatomical views. Table~\ref{tab:main}(b) additionally reports cluster-collapse counts and rates.

\item {\bf Experiments compute resources}
    \item[] Answer: \answerYes{}
    \item[] Justification: All experiments were conducted on a workstation with four NVIDIA RTX 5880 Ada GPUs, each with 48\,GB memory. Each individual training run used one GPU. The main experiments follow the 100-epoch configuration described in Section~\ref{subsec:exp-setup}. 

\item {\bf Code of ethics}
    \item[] Answer: \answerYes{}
    \item[] Justification: The research conforms with the NeurIPS Code of Ethics. We use only publicly available datasets (MNIST, Fashion-MNIST, OASIS) under their respective licenses, and the OASIS Brain MRI dataset is used in compliance with its data use agreement.

\item {\bf Broader impacts}
    \item[] Answer: \answerYes{}
    \item[] Justification: Our method enables more reliable identification of disease subtypes from medical imaging in the presence of noise, with potential positive impact on clinical analysis of disease heterogeneity (e.g., discovering rare subtypes that conventional models would merge). On the negative side, robust unsupervised clustering applied to medical data carries the standard risks of misinterpretation if used without clinical oversight, and like any method trained on retrospective cohorts may inherit demographic biases present in the source data; clinical deployment would require careful validation by domain experts and bias audits.

\item {\bf Safeguards}
    \item[] Answer: \answerNA{}
    \item[] Justification: The released assets consist of research code, documentation, and numerical experiment records for image prototype learning. No pretrained model weights, patient images, or subject-level clinical records are distributed. MRI data must be obtained separately through the dataset provider under its access conditions.

\item {\bf Licenses for existing assets}
    \item[] Answer: \answerYes{}
    \item[] Justification: All datasets and baseline implementations used in this paper are properly cited. MNIST~\citep{lecun1998gradient} is released under Creative Commons Attribution-Share Alike 3.0 license; Fashion-MNIST~\citep{xiao2017fashion} is released under MIT license; OASIS Brain MRI~\citep{lamontagne2019oasis} is used under its publicly available data use agreement. Baseline implementations of VaDE~\citep{jiang2017variational}, Contra-VaDE~\citep{li2021contrastive}, ClusterDDPM~\citep{yan2023clusterddpm}, $t^3$VAE~\citep{kim2024t3vae}, and the $t$-VaDE variants~\citep{kobayashi2019vade} are cited in Section~\ref{subsec:exp-setup}, and we re-implement them following the published descriptions to ensure architectural fairness.

\item {\bf New assets}
    \item[] Answer: \answerYes{}
    \item[] Justification: We release an MIT-licensed Python package implementing a reusable Gaussian--Gamma prior, a basic label-free Student's $t$ VAE, and the original multi-component models used by our method. The repository provides usage examples, training and evaluation scripts, configuration documentation, numerical audit results, and fresh reproduction reports.
The documentation distinguishes the basic single-component model from the paper's clustering models and states the scope and limitations of the available  reproduction results. The source code is available at
\url{https://github.com/xuzhikangnba/torch-tvae}.

\item {\bf Crowdsourcing and research with human subjects}
    \item[] Answer: \answerNA{}
    \item[] Justification: This paper does not involve crowdsourcing or new data collection from human subjects. The OASIS Brain MRI cohort is a pre-existing publicly released dataset; we do not perform new recruitment or data collection.

\item {\bf Institutional review board (IRB) approvals or equivalent for research with human subjects}
    \item[] Answer: \answerNA{}
    \item[] Justification: We use the OASIS Brain MRI dataset, which was collected under its own IRB-approved protocols by the original data providers, in accordance with their publicly available data use agreement. No new human subject research is conducted in this paper.

\item {\bf Declaration of LLM usage}
    \item[] Answer: \answerNA{}
    \item[] Justification: LLMs are not part of the core methodology of this work. They were used only as writing aids (grammar, phrasing) and do not impact the scientific rigor, originality, or experimental claims.

\end{enumerate}
\end{document}